%% file: main.tex
\documentclass[sigconf]{acmart}
\renewcommand\footnotetextcopyrightpermission[1]{} % removes footnote with conference information in first column
\AtBeginDocument{%
  \providecommand\BibTeX{{%
    \normalfont B\kern-0.5em{\scshape i\kern-0.25em b}\kern-0.8em\TeX}}}

\newbool{showComments}
\boolfalse{showComments}    % UNCOMMENT THIS BEFORE SUBMISSION
\ifbool{showComments}{%
\newcommand{\new}[1]{\textcolor{blue}{#1}}
}{
\newcommand{\new}[1]{\textcolor{black}{#1}}
}

\renewcommand\and{\\[\baselineskip]}
\begin{document}

% \author{Lorena Qendro^1, Jagmohan Chauhan^{1,2}, Alberto Gil C. P. Ramos^3, Cecilia Mascolo^1}
% \affiliation{%
%     \country{$^1$ University of Cambridge \cdot  $^2$ University of Southampton \cdot  $^3$ Samsung AI Center Cambridge%
%     }
% }

\author{Lorena Qendro}
\affiliation{%
  \institution{University of Cambridge}
%   \country{Country}
}
\author{Jagmohan Chauhan}
\affiliation{%
  \institution{University of Cambridge}
  \institution{University of Southampton}
%   \country{Country}
}
\author{Alberto Gil C. P. Ramos}
\affiliation{%
  \institution{Samsung AI Center Cambridge}
%   \country{Country}
}
\author{Cecilia Mascolo}
\affiliation{%
  \institution{University of Cambridge}
%   \country{Country}
}
\title{The Benefit of the Doubt: Uncertainty Aware Sensing for Edge Computing Platforms}

\renewcommand{\shortauthors}{Qendro et al.}

\input{abstract}

%   <concept_desc>Human-centered computing~Ubiquitous and mobile computing</concept_desc>
%   <concept_desc>Computing methodologies~Artificial intelligence; Supervised learning by classification.</concept_desc>

% \ccsdesc[500]{Human-centered computing~Ubiquitous and mobile computing}
% \ccsdesc[500]{Computing methodologies~Artificial intelligence; Supervised learning by classification.}

% %%
% %% Keywords. The author(s) should pick words that accurately describe
% %% the work being presented. Separate the keywords with commas.
% \keywords{Edge Platforms, Sensing, Bayesian Deep Learning, Uncertainty, Bayesian Approximation}

\maketitle

\pagestyle{plain}

\input{introduction}

\input{motivation}
\input{relatedwork}
\input{framework}
\input{evaluation}

\input{discussion}

\bibliographystyle{ACM-Reference-Format}
\bibliography{bibliography}

\end{document}

%% file: abstract.tex
\begin{abstract}
Neural networks (NNs) have drastically improved the performance of mobile and embedded applications but lack measures of ``reliability'' estimation that would enable reasoning over their predictions. Despite the vital importance, especially in areas of human well-being and health, state-of-the-art uncertainty estimation techniques are computationally expensive when applied to resource-constrained devices. We propose an efficient framework for predictive uncertainty estimation in NNs deployed on edge computing systems with no need for fine-tuning or re-training strategies. To meet the energy and latency requirements of these embedded platforms the framework is built from the ground up to provide predictive uncertainty based only on one forward pass and a negligible amount of additional matrix multiplications.
Our aim is to enable already trained deep learning models to generate uncertainty estimates on resource-limited devices at inference time focusing on classification tasks.
This framework is founded on theoretical developments casting dropout training as approximate inference in Bayesian NNs. Our novel layerwise distribution approximation to the convolution layer cascades through the network, providing uncertainty estimates in one single run \new {which ensures minimal overhead, especially compared with uncertainty techniques that require multiple forwards passes and an equal linear rise in energy and latency requirements making them unsuitable in practice.} We demonstrate that it yields better performance and flexibility over previous work based on multilayer perceptrons to obtain uncertainty estimates.
Our evaluation with mobile applications datasets on Nvidia Jetson TX2 and Nano shows that our approach not only obtains robust and accurate uncertainty estimations but also outperforms state-of-the-art methods in terms of systems performance, reducing energy consumption (up to 28--folds), keeping the memory overhead at a minimum while still improving accuracy (up to 16\%).
\end{abstract}

%% file: introduction.tex
\section{Introduction}
Deep learning models are becoming the de facto standard in mobile and embedded applications: examples include activity and context recognition~\cite{wang2018data, hsu2018video}, health and well-being monitoring~\cite{de2018heart, ryder2009ambulation,subasi2018iot}, and location prediction~\cite{ballinger2018deepheart,lu2012stresssense, ronao2016human,figo2010preprocessing}.
However, deep learning models are being also scrutinized due to their lack of interpretability. Understanding predictive uncertainty is important for all types of machine learning tasks but is key in situations when these are relied upon by the medical profession: examples of embedded systems in these domains are soaring, e.g for monitoring gait in Parkinson's disease patients~\cite{hammerla2016deep}, detection of cardiac arrest via audio from smart devices~\cite{chan2019contactless} or revealing sleep apnea~\cite{lin2017runtime} through sensing and audio applications.
 
\new{Probabilistic approaches exist to provide frameworks} for modeling uncertainty towards capturing the erroneous overconfident decisions. However, enabling \new{such approaches on deep neural network models bring significant challenges on embedded devices.}
Firstly, \new{the most popular approach to provide uncertainty estimates, namely} Bayesian Neural Networks (BNNs) require heavy computation. Although, recent efforts have been devoted to making them more efficient, their improvements are still not a good fit for mobile or embedded devices since they are based either on \textit{sampling}~\cite{gal2015bayesian,gal2016dropout, gal2016theoretically} or model \textit{ensembles}~\cite{lakshminarayanan2017simple}.
While sampling demands running a single stochastic neural network multiple times, ensemble methods require training and running multiple neural networks \new{which linearly increases latency if ran in sequence or memory if ran in parallel. Indeed,} 
these solutions are resource agnostic and would incur \new{unfeasible increases in power consumption, latency and memory requirements on many mobile devices with limited resources.
Secondly,  there is very limited work on \new{alternatives to BNNs} providing predictive uncertainty in embedded systems~\cite{yao2018rdeepsense, yao2018apdeepsense}  applicable only to multi layer perceptrons (MLPs). However, \new{as is well known in the aforementioned applications areas,}} using convolution neural networks (CNNs) instead leads to more accurate predictions than MLPs~\cite{hammerla2016deep, yao2017deepsense}. \new{Indeed,} the majority of modern embedded deep learning models do not rely solely on MLPs but are often a combination of different neural layers, CNNs and MLPs~\cite{yao2017deepsense, hammerla2016deep, mathur2019mic2mic}. As a consequence, these approaches \new{although suitable for embedded devices} are \new{not relevant for the types of deep learning models that are actually being deployed in practice}. Moreover, they focus mainly on regression tasks, leaving a considerable amount of questions on how they can be used in classification contexts.

In light of the highlighted challenges, we propose a framework that addresses these limitations by enabling \textit{predictive uncertainty} estimations 
for mobile and embedded applications and evaluating its efficiency on resource constrained devices.
Overall, we make the following contributions in this paper:

\begin{itemize}
    \item  We introduce an efficient framework that directly enables already trained deep learning models to generate uncertainty estimates with no need for re-training or fine-tuning. Its core is based on  theoretical developments casting dropout training as approximate inference in Bayesian Convolutional Neural Networks~\cite{gal2015bayesian}; we consider models that have been already trained with dropout as a regularization method.  This assumption is easily satisfiable, since most of the modern deep learning networks use dropout during training~\cite{yao2017deepsense, hammerla2016deep}. To achieve our goal in providing the uncertainty estimates, we propose an efficient layerwise distribution approximation, which transforms the single deterministic convolutional layer into a stochastic convolutional layer. Unlike previous methods that generate the prediction distribution via multiple runs~\cite{lakshminarayanan2017simple,gal2015bayesian,gal2016theoretically}, our layerwise distribution is propagated through the network in a cascaded manner \new{massively} reducing the computational complexity by allowing the model to produce uncertainty estimations in one single run. \new{This approach makes it possible therefore to enable predictive uncertainty on a much wider range of small devices where running uncertainty aware deep learning models would be impossible with traditional techniques.}
    
    \item Our approach focuses on classification tasks which makes obtaining uncertainty estimates challenging. Unlike regression, in a classification scenario, we cannot interpret the output distribution as the model prediction output. To solve this problem, we introduce an efficient way to marginalize over the final distribution to capture the predictive uncertainty and present the class accuracy. Moreover, our approach is able to offer the desired flexibility by enabling predictive uncertainty into CNNs which have better predictive power than MLPs. Combining CNNs with layerwise distribution approximations become a powerful tool to estimate uncertainty while offering higher accuracy compared to  the existing works which utilize MLP based models~\cite{yao2018rdeepsense, yao2018apdeepsense}.
    
    \item We evaluate our framework on the Nvidia Jetson TX2 and Nano embedded platforms on human activity recognition (HAR)  and audio sensing applications.   We compare our approach with the state-of-the-art Monte Carlo dropout~\cite{gal2016theoretically}, a fully connected network based approach~\cite{yao2018rdeepsense} as well as deep ensembles technique~\cite{lakshminarayanan2017simple}. For all approaches, we measure the resource  consumption (latency and energy) and model performance, such as the accuracy, and the quality of uncertainty estimations. Our approach can reduce inference and energy consumption by 8--fold to 28--fold, while obtaining robust and accurate uncertainty estimation. We also significantly improve the accuracy of the deep learning models, compared to previous work based on fully connected layered MLP models~\cite{yao2018apdeepsense} by a margin of at least 6\% to 16\% while being more cost-effective computationally. We make sure not too heavily contribute to the memory footprint by adding only a negligible runtime memory overhead (max 5\%) compared to the vanilla deep learning model and improving (by 30\%) on the MLP baseline. We show that our technique can smoothly run also on CPU only, allowing devices without GPU to still have fast and robust uncertainty aware predictions.
\end{itemize}

%% file: motivation.tex
\section{Motivation}
\label{motivation}
Limited previous work in mobile and embedded systems~\cite{yao2018apdeepsense, yao2018rdeepsense} has empirically studied ways to provide uncertainty estimations in deep learning models. These techniques mostly focus on regression, leaving the classification scenario relatively unexplored. Classification tasks make the highest percentage of mobile sensing applications~\cite{lane2015deepear, yao2017deepsense, hammerla2016deep, mathur2019mic2mic, de2018heart} but providing uncertainty estimations in the context of these types of data and resource constrained devices is still an open research area. Deterministic DNNs are trained to obtain maximum likelihood estimates and therefore do not consider uncertainty around the model parameters that leads to predictive uncertainty. They provide overconfident decisions as the softmax probability only captures the relative probability that the input is from a particular class compared to the other classes but not the overall model confidence.
\begin{figure}[t]
    \centering
    \begin{subfigure}[t]{0.45\columnwidth}
        \centering
        \includegraphics[width=\columnwidth]{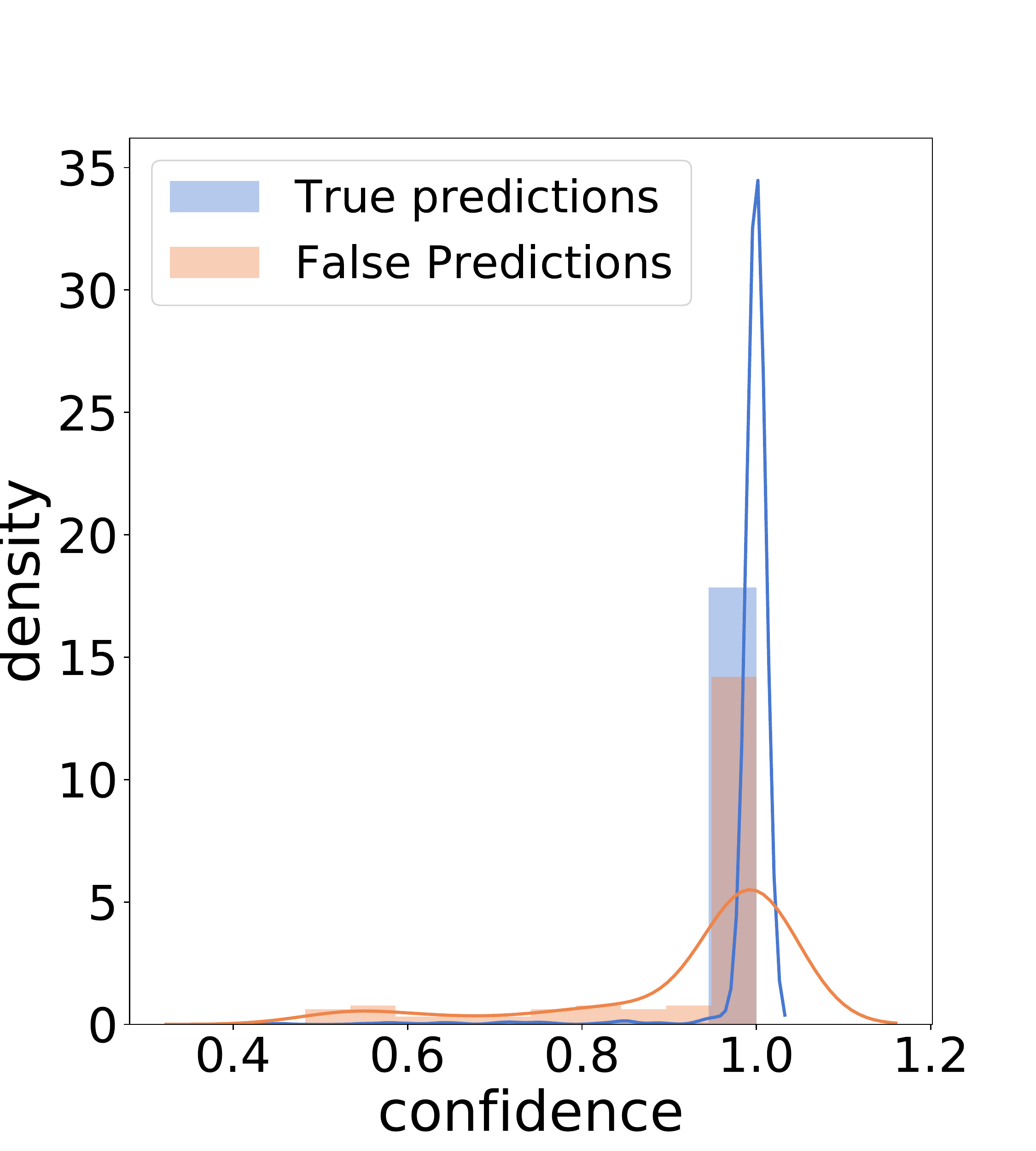}
        \caption{}
        \label{fig:hhar_confidence_dnn_motiv}
    \end{subfigure}
    \begin{subfigure}[t]{0.45\columnwidth}
        \centering
        \includegraphics[width=\columnwidth]{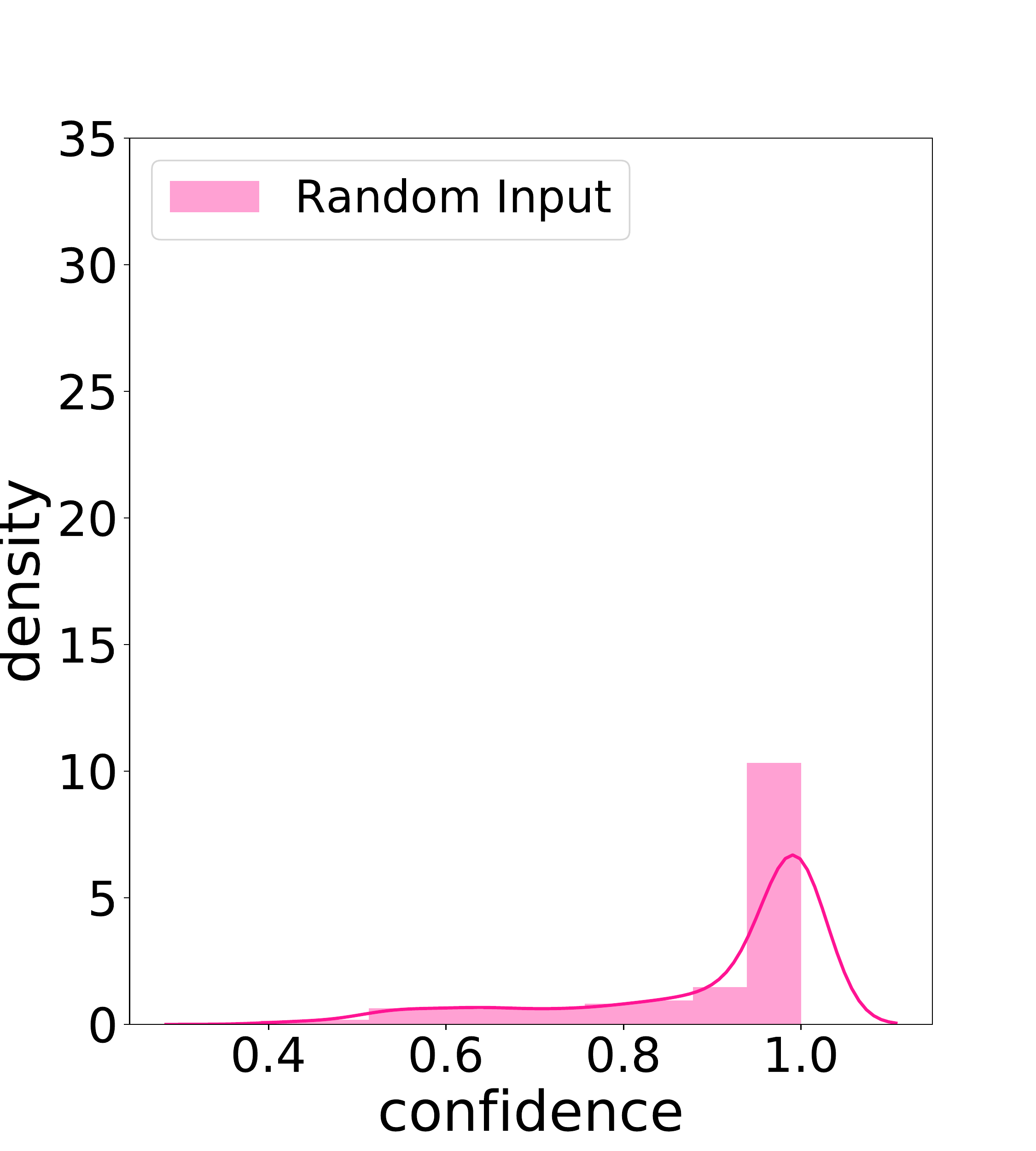}
        \caption{}
        \label{fig:hhar_confidence_random}
    \end{subfigure}
    \vspace{-0.10in}
  \caption{Density histogram of confidence measures using vanilla DNN. In (a), true (correct) and false (incorrect) predictions; in (b), random data.
  } 
  \label{fig:motiv_softmax}
    \vspace{-0.15in}
\end{figure}

To visualize the aforementioned issue, we  analyze a human activity recognition scenario. This experiment is performed on the Heterogeneous human activity recognition dataset (HHAR)~\cite{hhar} which consists of readings from two motions sensors (accelerometer and gyroscope) on nine users performing six activities (biking, sitting, standing, walking, climb stairs-up, and stairs-down). We choose a 5 layer network (4 convolutional layers and 1 fully connected) and evaluated the behavior of softmax on the test set as well as on random input data. In Figure~\ref{fig:hhar_confidence_dnn_motiv}, we can see the confidence measures for true (correct) and false (incorrect) predictions. A distribution skewed towards the right (near 1.0 on the x-axis) shows that the model has higher confidence in predictions than the distributions skewed towards left. As we can notice, this model gives high probability for both correct and incorrect predictions. Instead of the current outcome, we would like to see low confidence for the false predictions. An even more concerning observation can be noticed in ~\ref{fig:hhar_confidence_random} where we plot the confidence given by the same model on completely random data which shows that vanilla DNNs are overconfident even when presented with random data the model has not seen during training.
This limitation of conventional deep learning approaches motivates our work. We aim to provide accurate predictions while  understanding if the model is guessing at random or it is certain about the prediction. 

In addition, we want the deep learning models to run on resource-limited devices, therefore, to be latency and energy savvy. Our aim is to overcome the computation overhead of sampling-based Bayesian techniques and other non-Bayesian approaches like deep ensembles. \new{Running a single stochastic NN multiple times \new{for each prediction}, or \new{needing to retrain or fine-tune existing model(s), is not feasible for many edge platforms. To close this gap, we build a new framework that can enable uncertainty estimates for currently deployed models under the constraints that it must require only one forward pass for each prediction, no retraining or fine-tuning, and incur only a residual increase in latency and memory. What makes this possible is an approximation to the internal statistics of neural networks, that allows an approximate propagation of the signal and confidence through the network layers.} Its core is a layerwise distribution approximation which allows to create a stochastic convolution layer that enables} uncertainty estimates \new {to be approximately propagated from inputs to outputs in a single run, which incurs only negligible increased linear algebra cost}.

%% file: relatedwork.tex
\section{Related Work}
\label{related}

\new{The benefits that may result from providing uncertainty estimates for predictive models have long been recognized. Indeed, dating back to 1992, \new{seminal work in} \cite{mackay1992bayesian} shows several benefits of \new{stochastic approach\new{es} to neural network learning, \new{such as} naturally accounting for the model flexibility, aiding in comparison between different models, accurate calibration of predictive uncertainty, and robust\new{ness} to overfitting \new{to name a few. Naturally, the vast majority of this rich literature aims at small scale problems and precludes deployability considerations. More recently, there has been a significant scale up of the problems such techniques can be applied, but the subject matter of designing algorithms for uncertainty quantification on edge devices is in its infancy. In the following, we attempt to provide a brief overview of recent developments, and how this work extends those to a wider range of devices. }}} 

\subsection{Uncertainty Estimations} 

Modern research in Bayesian Neural Networks (BNN) relies on variational inference to approximate the intractable posterior of Bayesian inference~\cite{graves2011practical, blundell2015weight, welling2011bayesian, ahn2012bayesian,gong2018meta}.
Pioneering work from Gal et al.~\cite{gal2015bayesian, gal2016dropout, gal2016theoretically} introduced Monte Carlo dropout (MCDrop) interpreting dropout \new{used to train deterministic neural networks}~\cite{hinton2012improving, srivastava2014dropout} to approximately correspond to variational inference. MCDrop collects the results of stochastic forward passes through a model \new{with dropout enabled at inference time} and estimates the predictive uncertainty \new{via sampling-based techniques}. Deep Ensembles~\cite{lakshminarayanan2017simple}, instead, are a non-Bayesian way to get uncertainty estimations by training and running multiple DNNs.

All the work discussed so far focuses on \new{investigating} more accurate uncertainty estimates and do\new{es} not consider the system implications of mobile and embedded computing; as a consequence, the proposed methods \new{often \emph{i}) require training new models from scratch or retraining/fine-tuning existing models with a development cost that might inhibit their use and/or \emph{ii})} are computationally prohibitive \new{ i.e., require a linear increase in latency or memory due to multiple forward passes through a single model or one forward pass through several models}. \new{We on the other hand take a different approach and focus primarily on providing a simple and effective solution that enriches \new{existing} deep learning models with predictive uncertainty estimations \new{in a manner that does not require retraining/fine-tuning and that ensures a} latency\new{, memory} and energy consumption \new{in the same ballpark as the original model}}.

\subsection{Mobile Applications, Resource Constraints and Uncertainty}

Numerous works have investigated the use of deep neural networks for human activity recognition (HAR)~\cite{hammerla2016deep, radu2018multimodal, ravi2016deep, khan2018scaling, peng2018aroma} and audio sensing ~\cite{lane2015deepear, georgiev2017accelerating, georgiev2017low,bhattacharya2016sparsification,chauhan2018performance}. These applications need intelligence at the edge, and therefore, deal with constrained resources. Recently, traditional DNNs are modified to fit in memory, increase execution speed, and decrease energy demand to inference models on the edge \cite{Yang_2018_ECCV,cai2019once,lane2016deepx}. 
However, there is limited previous work that aims to enable uncertainty estimations on these models and platforms. In~\cite{yao2018rdeepsense,yao2018apdeepsense}, the authors propose an approximation to the output distribution using standard dropout~\cite{srivastava2014dropout}, which aims to reduce computation time and energy consumption. However, this work only applies to fully-connected NNs, leaving the challenges for more complex models, like Convolutional Neural Networks (CNNs), still to be addressed. Modern architectures are very rarely solely MLP based~\cite{yao2017deepsense, hammerla2016deep}. This suggests that extending CNNs to obtain uncertainty estimations makes it is possible to not only obtain higher accuracy but also more robust models in their predictive abilities. Moreover, the current works mainly focus on regression tasks, leaving the classification scenario relatively unexplored. Classification scenarios make the majority of mobile sensing applications, therefore, we intentionally focus on these tasks in our work.

Also, unlike previous works providing uncertainty estimation only in the last layer or after several runs, our approach provides flexibility through its layerwise approximation which captures the uncertain unobserved quantities at the layer level. This is useful in scenarios where models cannot execute all layers, possibly due to energy and latency constraints~\cite{lee2019neuro, montanari2020eperceptive} or would like to exit early in a densely layered model to save energy, while providing robust uncertainty estimates.

%% file: framework.tex
\section{Uncertainty Estimation Framework}
\label{framework}
 \begin{figure*}[t]
    \centering
    \includegraphics[width=1\textwidth]{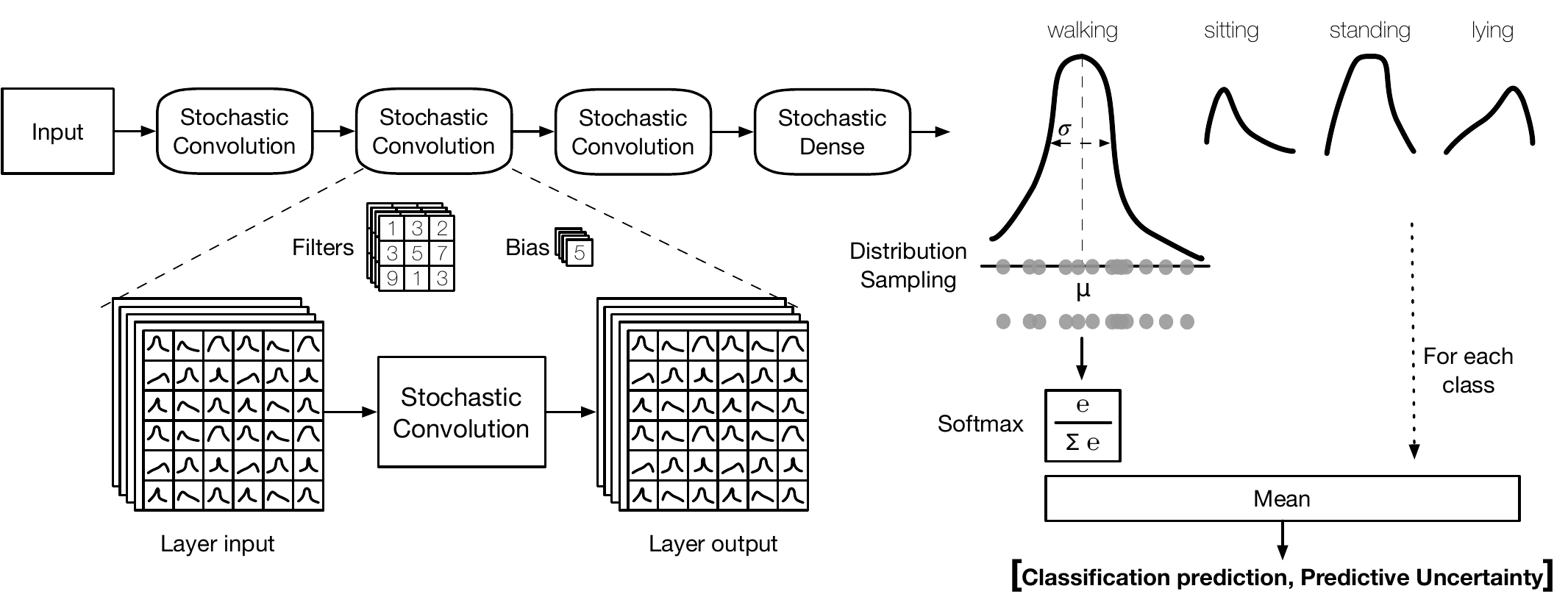}
    % \vspace{-0.15in}
    \caption{Overview of the framework: describing the composition of the stochastic convolution, it's propagation throughout the network, and finally the distribution sampling at the softmax layer.
    }
    % \vspace{-0.25in}
    \label{fig:overview}
\end{figure*}

We present a framework that enables pre-trained \new{deterministic} deep learning models to generate uncertainty estimates on resource-limited devices \new{\emph{i}) without increasing their latency, memory and energy profiles except for a minimal amount due to additional linear algebra and \emph{ii}) without any retraining/fine-tuning which could prevent deployability on the same devices. This is in stark contrast to existing techniques for providing uncertainty estimates which have a linear (often unaffordable) increase in operational costs due essentially to their reliance on multiple forward passes, and which may require training new models}.

\new{The building block of our approach is based on a layerwise distribution approximation. In particular,}
we enable the primary convolution operations performed inside a CNN to apply to the inputs described by probabilistic distributions (Figure~\ref{fig:overview}). Our approach does not require any additional re-training as long as the existing model has been trained with dropout regularization.

We define \textit{predictive uncertainty} as the probability of occurrence of the target variable conditioned on all the available information. We treat the deep learning model predictions as random variables~\cite{quinonero2005evaluating} framing the predictions as probability distributions instead of point estimates. 

 The framework is grounded on the fact that when using a neural network trained with dropout, the network does not have a deterministic structure anymore since it is described in part by random variables. One of our key contributions is to allow the basic convolution operations in the stochastic NNs generated by dropouts to not only output an expected value but a probability distribution of the output random variable.  However, computing the exact output distribution is not tractable. Therefore, we approximate it with a Multivariate Gaussian Distribution based on minimizing the Kullback-Leibler (KL) divergence between real and approximate distribution. 
 
 Since our novel convolution layer supports taking probabilistic distributions as inputs and generate distributions as outputs (Figure~\ref{fig:overview}), 
we integrate over these Gaussian distributions to ultimately  sample point values (unaries) and pass them through the softmax function.
At the end of this process, the classification model produces the class prediction and its predictive uncertainty based on the samples. This operation is extremely fast as we only run the model once, passing inputs to the models to get the output distribution. At this point, we only sample from the final distribution, which is a tiny fraction of the networks compute (see Section~\ref{energy}), and therefore, it does not significantly increase the model's inference time. 
 
 Figure \ref{fig:overview} provides a visual overview of the framework by describing the composition of the stochastic convolution, it's propagation throughout the network, and finally the distribution sampling at the SoftMax layer. Our approach uses the theoretical foundations of MCDrop~\cite{gal2015bayesian} casting dropout training as an implicit Bayesian approximation, however, it radically distinguishes from it as in our technique the distributions are embedded in the network and do not rely on multiple runs to produce them. In addition, we offer a novel mode to adapt the distribution output to predict the outcome in classification tasks while providing the desired predictive uncertainty estimations.

 \subsection{Background}
 \label{background}
 \subsubsection{Uncertainty definition}
 Uncertainty is manifested in various forms in mobile applications on the edge. It can be at the physical layer due to node mobility, network topology, routing and resource availability.~\cite{das2005coping} These factors together with the sensor measurements, calibration and software heterogeneity add variability to the system feeding uncertainty to it. When using deep learning the uncertainty induced by the model architecture and parameters is an additional factor that jeopardizes the trust in the prediction. Observed data can be consistent with many models, and therefore which model is appropriate, given the data, is uncertain. Similarly, predictions about future data and the future consequences of actions are uncertain. Probability theory provides a framework for modelling uncertainty~\cite{ghahramani2015probabilistic}.

In our work, we define \textit{predictive uncertainty} as the probability of occurrence of the target variable conditioned on all the available information. We treat the deep learning model predictions as random variables~\cite{quinonero2005evaluating}. This means that we will have the predictions as probability distributions instead of point estimates. These probability distributions are used to represent all the uncertain unobserved quantities in a model (including structural, parametric and noise-related). In Bayesian modelling, there are two main types of uncertainty that could be modelled~\cite{der2009aleatory}. \textit{Aleatoric} uncertainty captures the noise in the observations which might be sensor or motion noise. This noise gets translated into the uncertainty that cannot be reduced even if more data is collected. \textit{Epistemic} uncertainty (also called \textit{model uncertainty}), instead, represents the uncertainty in the model parameters and captures the ignorance about which model generated the collected data. \textit{Predictive uncertainty} incorporates both \textit{aleatoric} and \textit{epistemic} uncertainty. In this work, we are modelling both uncertainties in one single framework.

Predictive uncertainty provides a better interpretation of the inference because it can indicate if the deep learning model is certain about the prediction or is just guessing at random. If we consider interpretation as an understanding of all the elements that contributed to the prediction, uncertainty estimation is the first step to detecting the anomaly and acts as a trigger to perform further investigations.

\subsubsection{Dropout training (standard dropout)}
\label{drop}
\cite{srivastava2014dropout} proposes dropout as a regularization method to prevent over-fitting. The idea is to drop units from layers to avoid feature co-adaptation. For a \textit{fully connected (FC) neural network} the linear operation can be:

% \vspace{-0.10in}
{
\begin{align}
\begin{split}
\scriptstyle \textbf{y}^{(l)} &\scriptstyle= \scriptstyle \textbf{x}^{(l)}\textbf{W}^{(l)} + \textbf{b}^{(l)}
\\
\scriptstyle \textbf{x}^{(l+1)} &\scriptstyle= \scriptstyle f^{(l)}\big(\textbf{y}^{(l)}\big)
% \vspace{-0.10in}
\end{split}
\end{align}
}

where for each layer $\textit{l}$, $\textbf{x}^{(l)}$ and $\textbf{y}^{(l)}$ are the input and output of that layer, and  $\textbf{f}^{(l)}(\cdot)$ is a nonlinear activation function. $\textbf{W}^{(l)}$ is the weight matrix of \textit{l} with dimensions $\textit{K}^{(l)}$ x $\textit{K}^{(l-1)}$ and $\textbf{b}^{(l)}$ is the bias vector of dimensions $\textit{K}^{(l)}$. Using dropout at the $l^{th}$ layer is mathematically equivalent to setting the rows of the weight matrix $\textbf{W}^{(l)}$ for that layer to zero. The FC layer can, therefore, be represented with dropout:

% \vspace{-0.10in}
{
\begin{align}
\begin{split}
\scriptstyle \textbf{z}_{[i]}^{(l)} \scriptstyle &\sim \scriptstyle \mathrm{Bernoulli}\big(\cdot|\textbf{p}_{[i]}^{(l)}\big)
\\
\scriptstyle \tilde{\textbf{W}}^{(l)} \scriptstyle &= \scriptstyle \mathrm{diag}\big(\textbf{z}^{(l)}\big)\textbf{W}^{(l)}
\\
\scriptstyle \textbf{y}^{(l)} \scriptstyle &= \scriptstyle \textbf{x}^{(l)}\tilde{\textbf{W}}^{(l)} + \textbf{b}^{(l)}
\\
\scriptstyle \textbf{x}^{(l+1)} \scriptstyle &= \scriptstyle f^{(l)}\big(\textbf{y}^{(l)}\big)
\end{split}
\end{align}
}

Here $\textbf{z}_{[i]}^{(l)}$ are Bernoulli distributed random variables with some probabilities $\textbf{p}_{[i]}^{(l)}$. The $\mathrm{diag}(\cdot)$ maps vectors to diagonal matrices.

The described dropout operations convert a deterministic NN with parameters $\textbf{W}^{(l)}$ into a random Bayesian neural network with random variables $\tilde{\textbf{W}}^{(l)}$, which equates to a NN with a statistical model without using the Bayesian approach explicitly.

\begin{figure*}[t]
    \centering
    \includegraphics[width=0.8\textwidth]{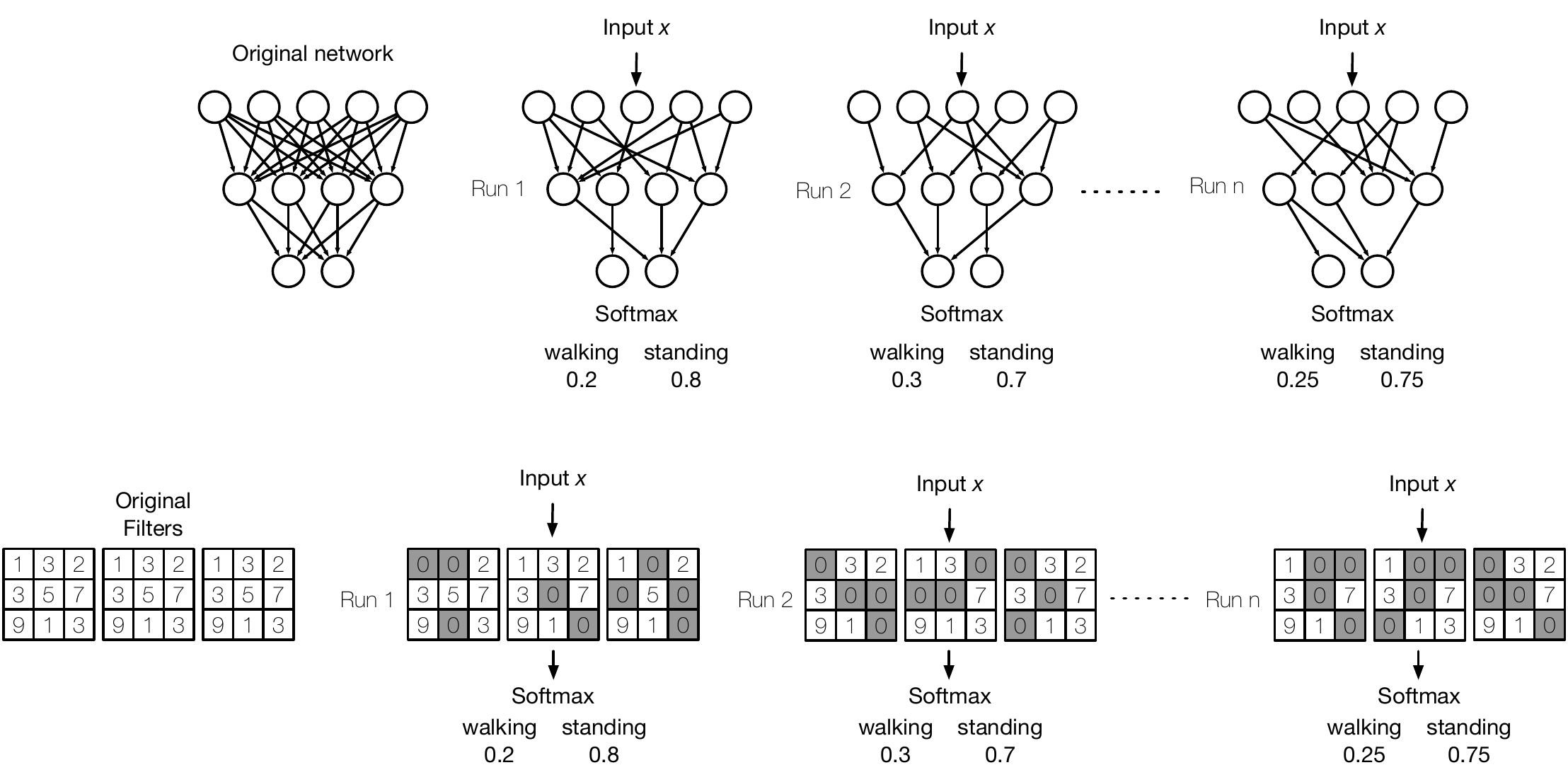}
    \caption{Monte Carlo dropout performed in fully-connected and convolutional neural networks. Keeping dropout during inference creates an implicit ensemble of models. As shown it requires running the same network with different dropout masks in order to provide uncertainty estimations.
    }
    % \vspace{-0.25in}
    \label{fig:montecarlo}
\end{figure*}

\subsubsection{Dropout as Bayesian approximation}
\label{mcdrop}

\cite{gal2015bayesian} proves the equivalence between dropout training in a neural network and approximate inference in a deep Gaussian process (GP). Differently from a non-probabilistic NN, a deep GP is a powerful statistical tool that allows to model distributions over functions. This means that to formulate the neural network layer as a Gaussian process we would define its covariance function

% \vspace{-0.07in}
{
\begin{align}
    \scriptstyle \textbf{K}(\textbf{x}, {\textbf{x}}') =& \scriptstyle \int p\big(\textbf{W}^{(l)}\big)f^{(l)}\big(\textbf{x}\textbf{W}^{(l)} + \textbf{b}^{(l)}\big)f^{(l)}\big({\textbf{x}}'\textbf{W}^{(l)} + \textbf{b}^{(l)}\big)d\textbf{W}^{(l)},
\end{align}
}

with an element-wise linearity $f^{(l)}(\cdot)$ and distribution $p(\textbf{W}^{(l)})$. If we consider now a deep GP with \textit{L} layers and covariance function $ \textbf{K}(\textbf{x}, {\textbf{x}}')$, it can be approximated by setting a variational distribution over each element of a spectral decomposition of the covariance function of the GP. This spectral decomposition maps each layer of the deep GP to a layer of hidden units in the NN. For an \textit{L}-layer neural network, this suggests that we can feed the output of one GP to the covariance of the next GP matching a deep GP~\cite{damianou2013deep}. Hence, the final predictive distribution can be formulated as

% \vspace{-0.10in}
{
\begin{align}
    \scriptstyle p(\textbf{y}|\textbf{x}, \textbf{X},\textbf{Y}) = \scriptstyle \int p(\textbf{y}|\textbf{x}, \mathcal{W})p(\mathcal{W}|\textbf{X},\textbf{Y})d\mathcal{W},
\end{align}
}

where $p(\textbf{y}|\textbf{x}, \mathcal{W})$ is the whole Bayesian NN posterior with random variables $\mathcal{W} = \{p\big(\textbf{W}^{(l)}\big)\}$. To infer the predictive distribution estimation $p(\textbf{y}|\textbf{x}, \textbf{X},\textbf{Y})$ we need to calculate $p(\textbf{y}|\textbf{x}, \mathcal{W})$ which is intractable. To address this, Gal et al.~\cite{gal2015bayesian} proposed to use $q\big(\tilde{\textbf{W}}^{(l)}\big)$, a distribution over the weight matrices as follows: 

% \vspace{-0.10in}
{
\begin{align}
\begin{split}
\scriptstyle \textbf{z}_{[i]}^{(l)} &\sim \scriptstyle \mathrm{Bernoulli}\big(\cdot|\textbf{p}_{[i]}^{(l)}\big)
\\
\scriptstyle q\big(\tilde{\textbf{W}}^{(l)}\big) &= \scriptstyle \mathrm{diag}\big(\textbf{z}^{(l)}\big)\textbf{W}^{(l)}
\end{split}
\end{align}
}

The true posterior distribution is, therefore, approximated by the variational distribution $q\big(\tilde{\textbf{W}}^{(l)}\big)$ where $\tilde{\textbf{W}}^{(l)}$ represents the random variables used in dropout operations as described in (2). Finally, to approximate the predictive distribution $p(\textbf{y}|\textbf{x}, \textbf{X},\textbf{Y})$ they perform Monte Carlo (MC) sampling of the random variables $\mathcal{W}$,

% \vspace{-0.10in}
{
\begin{align}
    \scriptstyle q(\textbf{y}|\textbf{x}) = \scriptstyle \frac{1}{T} \sum_{t=1}^{T}q(\textbf{y}|\textbf{x}, \mathcal{W}_t),
\end{align}
}

where T is the number of MC samples. This method is called Monte Carlo Dropout (MCDrop) and is equivalent to performing  T stochastic passes. 

In the same way as using dropout on the FC layer, MCDrop can be applied to the individual convolutions in \textit{convolution neural networks}~\cite{gal2015bayesian}. The convolution process is an iterative process that takes a sequence of steps in order to compute all elements in the input layer. Similarly to the FC layers, we can sample Bernoulli random variables $\textbf{z}_{i,j,k}$ and apply them as masks to the weight matrix $\textbf{W}_i \cdot diag([\textbf{z}_i,j,k])$ which is equivalent to setting weights to 0 for different elements of the input.

Figure \ref{fig:montecarlo} illustrates how MC dropout is implemented. This technique relies on MC sampling and requires the whole network to run multiple times. Performing the multiple forward passes creates an implicit ensemble of models that differ to one other due to changes in the dropout masks. This implies that different runs will have different nodes and connections and provide the desired stochasticity in the Bayesian deep learning model. The predictive distribution is, therefore, not produced by explicitly placing distribution over the layers but by running multiple stochastic forward passes with dropout activated also during inference. These runs are then averaged and give an indication of the predictive uncertainty provided by the variance in the output, distribution variance in the regression context and SoftMax variance in the classification. Although MCDrop is a step forward towards optimizing and accelerating Bayesian deep learning techniques, it is not enough for running on edge platforms.

In this paper, we propose a novel solution, based on the theoretical foundations described in (3), (4), and (5), with the benefit of providing predictive uncertainties with solely one forward pass and test it on mobile applications running on embedded platforms.

\subsection{Our Approach to Efficient Uncertainty Estimation}
\label{our-approach}
Our approach replaces the slow and computationally intense sampling process with a resource-friendly method by enabling convolution neural networks to output predictive uncertainty alongside the prediction. 

We firstly propose a layerwise distribution approximation which allows to incorporate the distribution at the layer level and propagate it throughout the network. We choose this distribution to be a multivariate Gaussian distribution based on the equivalence of GPs and neural networks trained with dropout. In Section \ref{mcdrop}, (3), (4), and (5) provide the theoretical proof that a deep neural network trained with dropout could be represented by a deep Gaussian process. In (3) we can see how the GP would be represented at the layer level and (4) formulates it for the whole deep learning network by feeding the output of one Gaussian process to the covariance of the next GP, and therefore matching a deep GP. Accordingly, the internal layer of the deep neural network can be represented by the internal representations of deep Gaussian processes. A GP is a stochastic process such that every finite collection of those random variables has a multivariate Gaussian distribution.

Therefore, we \new{initially considered our layerwise distribution approximation to be represented by multivariate normal distributions. We found, however, that this approximation was not enough for avoiding multiple forward passes, and therefore explored an additional approximation of Gaussian distributions with diagonal covariance matrices.} \new{Indeed, by modeling inputs and outputs of each layer as approximately following Normal distributions with diagonal matrices, it is possible to compute the outputs mean and standard deviation in closed-form from those of the inputs as well as the layer operations for deterministic models trained with dropout, without requiring any retraining.}

Our approach is based on \new{an approximation to the internal statistics of neural networks, that permits an approximate propagation of the signal and confidence through the network layers, in a manner that is applicable to convolution and dense layers.}

To start, we enable the basic operations in convolution neural networks to output the expected value and a probability distribution of the output random variable. 
Computing the exact output distribution is intractable; consequently, we approximate it with the multivariate Gaussian distribution.  This approximation is based on minimizing the Kullback-Leibler (KL) divergence between the exact and approximate distributions.

% \vspace{-0.10in}
{
\begin{align}
\begin{split}
 \scriptstyle \min _{q} \mathrm{KL}(p(x) \| q(x)) =& \scriptstyle \min _{q} \int p(x) \log \left(\frac{p(x)}{q(x)}\right) d x \\=& \scriptstyle \min _{\mu, \sigma^{2}} -\int p(x) \log \mathcal{N} \left(x|\mu, \sigma^{2}\right) d x \\=& \scriptstyle \min _{\mu, \sigma^{2}} \frac{\log (\sigma^{2})}{2}+\frac{\int_{x} p(x)(x-\mu)^{2} d x}{2 \sigma^{2}}
 \end{split}
\end{align}
}

Where $p(x)$ is the exact output distribution and $q(s)\sim \mathcal{N} \left(x|\mu, \sigma^{2}\right)$ is the approximate output distribution. 
To obtain the optimal approximate output distribution we take the derivative over the $\mu$ and $\sigma^{2}$, therefore, the approximation can be represented as matching the mean and variance  between the two distributions.

% \vspace{-0.10in}
{
\begin{align}
\begin{split}
 \scriptstyle\mu =& \scriptstyle\int p(x)xdx \\ \scriptstyle \sigma ^{2} =& \scriptstyle \int p(x)(x-\mu)^{2}dx
\end{split}
\end{align}
}

We reformulate the convolution operation with dropout as follows. The input to the layer is represented by \(\textbf{x} \in \mathbb{R}^{(H,W,C)}\) with height \(H\), width \(W\) and \(C\) channels. Let \(\textbf{w} \in \mathbb{R}^{(h,w,c,f)}\) be the weight matrix with height \(h\), width \(w\), \(c\) channels and \(f\) filters, and \(\textbf{b}\in\mathbb{R}^{(f)}\) the bias vector. Consequently, the output will be represented as \(\textbf{y}_{[:,:,k]} := \textbf{x} * \textbf{w}_{[:,:,:,k]} + \textbf{b}_{[k]}\) for \(\textbf{y} \in \mathbb{R}^{(H-h+1,W-w+1,f)}\) and for \(k = 1,\ldots,f\) where $\odot$ represents the element-wise operation and $*$ the convolution operation. Formally,

% \vspace{-0.10in}
{
\begin{align*}
\begin{split}
    \scriptscriptstyle \textbf{y}_{[i,j,k]} \scriptscriptstyle:= \scriptscriptstyle \mathrm{vec}\big(\textbf{x}_{[i+[0:h],j+[0:w],:]} \odot \textbf{z}_{[i+[0:h],j+[0:w],:]}\big)^\top \mathrm{vec}\big(\textbf{w}_{[:,:,:,k]}\big) \scriptscriptstyle \scriptscriptstyle+& \scriptscriptstyle \textbf{b}_{[k]}
\end{split}
\end{align*}
}

where

% \vspace{-0.10in}
{
\begin{align}
\begin{split}
    \scriptstyle \textbf{z}_{[i+[0:h],j+[0:w],l]} &\sim \scriptstyle \mathrm{Bernoulli}\big(\cdot|\textbf{p}_{[i+[0:h],j+[0:w],l]}\big)\\
    \scriptstyle \textbf{x}_{[i+[0:h],j+[0:w],l]} &\sim \scriptstyle \mathcal{N}\big(\cdot|\boldsymbol{\mu}_{[i+[0:h],j+[0:w],:]},\boldsymbol{\sigma}_{[i+[0:h],j+[0:w],l]}^2\big)
\end{split}
\end{align}
}

According to (7), we need to calculate the mean and the variance of the output distribution $p(\textbf{y}_{[i,j,k]})$. Given that the Bernoulli variables \textbf{z} and the input Gaussian variables \textbf{x}, as shown in (9), are independent random variables, we can have the mean of the output as follows:

{
\begin{align}
\begin{split}
    \scriptstyle \mathbb{E}\big[\textbf{y}_{[i,j,k]}\big] &=
    \scriptstyle \mathbb{E}\left[\sum_{u=i}^{i+h-1} \sum_{v=j}^{j+w-1} \sum_{a=1}^{c} \textbf{x}_{[u,v,a]} \textbf{z}_{[u,v,a]} \textbf{w}_{[u-i,v-j,a]}\right]  + \scriptstyle \textbf{b}_{[k]}\nonumber\\
    &= \scriptstyle \sum_{u=i}^{i+h-1} \sum_{v=j}^{j+w-1} \sum_{a=1}^{c} \mathbb{E}\big[\textbf{x}_{[u,v,a]}\big] \mathbb{E}\big[\textbf{z}_{[u,v,a]}\big] \mathbb{E}\big[\textbf{w}_{[u-i,v-j,a]}\big] + \scriptstyle \textbf{b}_{[k]}\nonumber\\
    &= \scriptstyle \sum_{u=i}^{i+h-1} \sum_{v=j}^{j+w-1} \sum_{a=1}^{c} \boldsymbol{\mu}_{[u,v,a]} \textbf{p}_{[u,v,a]} \textbf{w}_{[u-i,w-j,a]} + \scriptstyle \textbf{b}_{[k]}
    % \label{eq: expectation}
\end{split}
\end{align}
}

Since $\textbf{x}_{[u,v,a]} \textbf{z}_{[u,v,a]} \textbf{w}_{[u-i,v-j,a]}$ are independent variables, 
we can measure the variance as:

\begin{align}
 \begin{split}
    \scriptstyle \mathrm{Var}\big[\textbf{y}_{[i,j,k]}\big] \scriptstyle &= 
    \scriptstyle \mathrm{Var}\left[\sum_{u=i}^{i+h-1} \sum_{v=j}^{j+w-1} \sum_{a=1}^{c} \textbf{x}_{[u,v,a]} \textbf{z}_{[u,v,a]} \textbf{w}_{[u-i,v-j,a]} + \scriptstyle\textbf{b}_{[k]}\right]\nonumber\\
    &= \scriptstyle \sum_{u=i}^{i+h-1} \sum_{v=j}^{j+w-1} \sum_{a=1}^{c} \mathrm{Var}\big[\textbf{x}_{[u,v,a]} \textbf{z}_{[u,v,a]} \textbf{w}_{[u-i,v-j,a]} + \textbf{b}_{[k]}\big]\nonumber\\
    &= \scriptstyle \sum_{u=i}^{i+h-1} \sum_{v=j}^{j+w-1} \sum_{a=1}^{c} \mathbb{E}\big[\big(\textbf{x}_{[u,v,a]} \textbf{z}_{[u,v,a]} \textbf{w}_{[u-i,v-j,a]} + \textbf{b}_{[k]}\big)^2\big] \\ &- \scriptstyle \big(\mathbb{E}\big[\textbf{x}_{[u,v,a]} \textbf{z}_{[u,v,a]} \textbf{w}_{[u-i,v-j,a]} + \textbf{b}_{[k]}\big]\big)^2\nonumber\\ 
    % =&  \scriptstyle \sum_{u=1}^{i+h-1} \sum_{v=j}^{j+w-1} \sum_{a=1}^{c} \mathbb{E}[\textbf{x}_{u,v,a}^2] \mathbb{E}[\textbf{z}_{u,v,a}^2] \mathbb{E}[\textbf{w}_{u-i,v-j,a}^2] - \mathbb{E}^2[\textbf{x}_{u,v,a}] \mathbb{E}^2[\textbf{z}_{u,v,a}] \mathbb{E}^2[\textbf{w}_{u-i,v-j,a}]\\
    % =& \scriptstyle \sum_{u=1}^{i+h-1} \sum_{v=j}^{j+w-1} \sum_{a=1}^{c} (\boldsymbol{\mu}_{u,v,a}^2 + \boldsymbol{\sigma}_{u,v,a}^2) \textbf{p}_{u,v,a} \textbf{w}_{u-i,w-j,a}^2 - \boldsymbol{\mu}_{u,v,a}^2 \textbf{p}_{u,v,a}^2 \textbf{w}_{u-i,w-j,a}^2\\
    % =& \scriptstyle \sum_{u=1}^{i+h-1} \sum_{v=j}^{j+w-1} \sum_{a=1}^{c} \mathbb{E}[\textbf{x}_{u,v,a}^2] \mathbb{E}[\textbf{z}_{u,v,a}^2] \mathbb{E}[\textbf{w}_{u-i,v-j,a}^2] - \mathbb{E}^2[\textbf{x}_{u,v,a}] \mathbb{E}^2[\textbf{z}_{u,v,a}] \mathbb{E}^2[\textbf{w}_{u-i,v-j,a}]\\
    \scriptstyle &= \scriptstyle \sum_{u=i}^{i+h-1} \sum_{v=j}^{j+w-1} \sum_{a=1}^{c} \big(\big(\boldsymbol{\mu}_{[u,v,a]}^2 + \boldsymbol{\sigma}_{[u,v,a]}^2\big) \textbf{p}_{[u,v,a]} \\ &- \scriptstyle \boldsymbol{\mu}_{[u,v,a]}^2 \textbf{p}_{[u,v,a]}^2\big) \textbf{w}_{[u-i,w-j,a]}^2
    \label{eq: variance}
 \end{split}
\end{align}

We can further represent the operations in a simple way and efficiently compute the output distribution, namely

% \vspace{-0.10in}
{
\begin{align}
\begin{split}
\scriptstyle \mathbb{E}[\boldsymbol{y}] \scriptstyle &= \scriptstyle (\boldsymbol{\mu}\odot\boldsymbol{p})*\boldsymbol{w} + \boldsymbol{b}\\
\scriptstyle \mathrm{Var}[\boldsymbol{y}] \scriptstyle &= \scriptstyle (((\boldsymbol{\mu}\odot \boldsymbol{\mu}) + (\boldsymbol{\sigma}\odot \boldsymbol{\sigma}))\odot \boldsymbol{p} - (\boldsymbol{\mu}\odot \boldsymbol{\mu}) \odot (\boldsymbol{p}\odot \boldsymbol{p})) * (\boldsymbol{w}\odot \boldsymbol{w})\\
    \scriptstyle \boldsymbol{y}_{[i,j,k]} \scriptstyle &\sim \scriptstyle \mathcal{N}\big(\cdot|\mathbb{E}\big[\boldsymbol{y}_{[i,j,k]}\big], \mathrm{Var}\big[\boldsymbol{y}_{[i,j,k]}\big]\big)
\end{split}
\end{align}
}

We have provided a mathematically grounded proof on how to calculate the mean and the variance of the output at each convolution layer on networks that have been trained with dropout. This means that we can have CNNs taking probabilistic distributions as inputs and generate distributions as outputs. Hence, avoiding the need for the computationally costly sampling to from these distributions. 

\new{Given our approximation of modeling the inputs and the outputs as Gaussian distributions with diagonal covariance matrix, what is required now is to compute the mean and variance of the activation function that follows the linear mapping, which can then be plugged into the Gaussian model. Towards this end,
the mean and variance}

\begin{align}
\begin{split}
    \scriptstyle \boldsymbol{\mu}_{\boldsymbol{y}} = \mathbb{E}[\boldsymbol{y}]\\
    \scriptstyle \boldsymbol{\sigma}_{\boldsymbol{y}}^2 = \mathbb{E}[\boldsymbol{y} - \boldsymbol{\mu}_{\boldsymbol{y}}]^2
\end{split}
\end{align}

\new{can be represented as a sum of expectations of the output activations with respect to the Gaussian input distribution, over the compact intervals where the activation function is linear.
This computation can be done in closed-form via $\mathrm{erf(\cdot)}$ for any piece-wise linear activation
as demonstrated in~\cite{yao2018apdeepsense}, and in particular for the ReLU activation used in our work.}

\begin {table*}[ht]
\caption {Statistical information of the publicly available datasets used for evaluation. \textit{Heterogeneous} human activity recognition (HHAR)~\cite{stisen2015smart}, \textit{Opportunity} ~\cite{chavarriaga2013opportunity} and \textit{Speech Commands} ~\cite{speech-commands} dataset.}
\begin{center}
{%
 \begin{tabular}{c | c | c|p{7 cm} | c} 
 \hline
 Dataset & Training size & Testing size & {Output} & One-hot Encoding\\ [0.5ex] 
 \hline\hline
 HHAR & 28,314& 1,686 & Sit, Stand, Walk, Bike, StairUp, StairDown& \checkmark\\
 \hline
 Opportunity & 26,908 & 6,287 & Stand, Walk, Sit and Lie& \checkmark\\
 \hline
 Speech Commands & 1,668 & 562 & Yes, No, Up, Down, Left, Right, On, Off, Stop, Go& \checkmark\\
 \hline
\end{tabular}}
\label{tab:data}
\end{center}
\end{table*}

Secondly, 
we propose an efficient way to exploit the output distribution of our stochastic neural network to provide a classification prediction and the predictive uncertainty measure. To this aim, we marginalise these Gaussian distributions in the logit space. 
We sample unaries (single elements) from the output distribution and then pass the point values from this distribution to the softmax function.

% \vspace{-0.10in}
{
\begin{equation}
\begin{split}
    \scriptstyle \boldsymbol{y} \scriptstyle \sim \scriptstyle \mathcal{N}\big(\cdot|\mathbb{E}[\boldsymbol{y}], \mathrm{Var}[\boldsymbol{y}]\big)\\
    \scriptstyle p(y = c| \textbf{x}, \textbf{X}, \textbf{Y}) \scriptstyle \approx \scriptstyle \frac{1}{S}\sum_{s=1}^{S} Softmax(\textbf{y}_s)\\
    \scriptstyle H(\textbf{y}|\textbf{x}, D) \scriptstyle = \scriptstyle - \sum_{i=1}^{S}p_{c\mu} * log p_{c\mu},
    \end{split}
\end{equation}
}

where $\textbf{y}$ is the output distribution of the model. Therefore, the prediction can be considered as the mean of the categorical distribution obtained from sampling $S$ single values from the Gaussian output distribution and then squashed with the softmax function to obtain the probability vector and the predictive entropy. The sampling operation from the output distribution is extremely fast as we only run the model once, passing inputs to the models to get the output logits. At this point, we only sample from the logits, which is a tiny fraction of the networks compute, and as we can see in Section \ref{energy}, it does not significantly increase the model's inference time.

In conclusion, we add a layerwise approximation to the convolutional layers which is propagated throughout the network to produce a probability distribution in output. With the approximation presented in (10) and the output distribution sampling in (16) we can now enable classification models to output predictive uncertainties alongside the class inference in one single run.

%% file: evaluation.tex
\section{Evaluation}
\label{eval}

To evaluate the performance of our approach, we build a five-layer deep neural network composed of four 2D convolutional layers with a ReLU activation function and one fully-connected output layer. The choice of the architecture was made to make a fair comparison with the other baselines which mainly relied on five-layer deep networks~\cite{yao2018apdeepsense,yao2018rdeepsense,lakshminarayanan2017simple}.

During training, the model is optimized by ADAM~\cite{kingma2014adam} with a learning rate of 1e-4. We add dropout of 0.5 (default) at each internal layer to stabilize training, avoid overfitting, and fulfill our requirement of having a model trained with dropout regularization. We use cross entropy as the loss function and a batch size of 64. For all the datasets, we use 5\% of the training set  for validation and hyper-parameter tuning. We employ the described architecture for all the datasets.

During inference, we enable our layerwise approximation to all layers and propagate it all the way to the output layer (Figure~\ref{fig:overview}). This architecture allows to have the output of the model represented as  mean and variance of the output distribution. However, since we are dealing with classification tasks, we sample unaries from the output distribution and pass them through the softmax function, as explained in Section~\ref{our-approach}. We have two inputs to the first layer of the NN: the sample we need to do inference on and the standard deviation of the data calculated on the training set. We use this as a prior to feed on the stochastic network.

\subsection{Datasets}
\textit{Heterogeneous} human activity recognition dataset (HHAR)~\cite{stisen2015smart} contains readings from two motion sensors (accelerometer and gyroscope). The data is gathered from nine users performing six activities (biking, sitting, standing, walking, climb stairs-up, and climb stairs-down) with six types of mobile devices. Conforming to the description on~\cite{yao2018apdeepsense}, we segment raw measurements into five seconds samples and take Fourier transformation on these samples as the input data. Each sample is further divided into time intervals of length 0.25s.

\textit{Opportunity} dataset~\cite{chavarriaga2013opportunity} consists of data from multiple accelerometers and gyroscopes placed on participants' body at different locations
such as arms, back, and feet. 
We used three devices on a hip, a left lower arm, and a right shoe by default and target to detect the mode of locomotion: stand, walk, sit and lie. In total, for all users and all recordings, the dataset consists of 3,653 modes of locomotion instances of variable duration (between ~0.2 and ~280 seconds). Following the preprocessing proposed by~\cite{hammerla2016deep}, we use run 2 from user 1, runs 4 and 5 from user 2 and 3 in our test set. The remaining data is used for training. For frame-by-frame analysis, we created sliding windows of duration 1 second with 50\% overlap.

For audio sensing, we use the \textit{Speech Commands} dataset~\cite{speech-commands} and the suggested preprocessing by~\cite{mathur2019mic2mic}. We train our network with the portion of the dataset that consists of 2,250 one-second long speech files belonging to 10 keyword classes (yes, no, up, down, left, right, on, off, stop, and go). In this task, the goal is to identify the presence of a certain keyword class in a given speech segment. This 10-class dataset was then randomly split into training (75\%) and test (25\%) class-balanced subsets to make sure we get the same amount of data for each class. The input to the model is a two-dimensional tensor extracted from the keyword recording, consisting of time frames and 24 MFCC features.

All three of our datasets were collected in a controlled environment, therefore, we augmented real-life noise to the datasets in a principled manner to include real-world variability. Data augmentation can encode prior knowledge on the data, result in more robust models, and provide more resources to the deep learning platform. For both HAR datasets, we used the data augmentation techniques proposed in~\cite{um2017data} which consist of seven variations for IMU data on wearables: rotation, permutation, time-warping, scaling, magnitude-warping, jittering, and cropping. 
For example, permutation consists of randomly perturbing the temporal location of events within the same time window. Rotation, instead, consists of enriching the data to match different sensor placements like an upside-down position. 
These alterations allow us to consider real-life noise which cannot be witnessed in the data collected in a lab-controlled setting, e.g., rotating of a smartwatch. 
We randomly chose a period between 20 seconds and two minutes and applied a randomly selected augmentation method; the intervals between noise periods were randomly selected between two and five minutes.
For audio, we sampled examples of ambient noise from a publicly available environment sound classification dataset~\cite{piczak2015esc} and added them to the audio dataset. We assume that only one single type of noise is present at a given time, and each noise lasts between 15 and 60 seconds.

\begin {table*}[ht]
% \vspace{-0.10in}
\caption {Accuracy, F1 Score, and Negative Log-Likelihood (NLL) for HHAR, Opportunity (Opp.), and Speech Command (SC) dataset. MCDrop-\textit{k} represents MCDrop with \textit{k} (3, 5, 10, and 30) runs (forward passes), instead DeepEnsemble-\textit{k} represents an ensemble of \textit{k} (3, 5, and 10) individual NNs.
NLL measures the correspondence between the ground truth values and their predicted distributions. Lower NLL means higher correspondence. Bold entries indicate best performance.} \label{tab:acc1}
% \vspace{-0.10in}
\begin{center}
\resizebox{1\textwidth}{!}{\begin{tabular}{c|c | c | c | c|c| c | c | c | c|c} 
 \hline
 \textbf{Model}&\textbf{Ours}&\textbf{BackboneDNN} &	 \textbf{ApDeepSense}	 & \textbf{MCDrop--3} 	 &	 \textbf{MCDrop--5} 	 &	 \textbf{MCDrop--10} 	 &	 \textbf{MCDrop--30} 	 &	 \textbf{Ensemble--3} 	 &	 \textbf{Ensemble--5} 	 &	 \textbf{Ensemble--10}\\ [0.5ex]  \hline\hline
 \textbf{Accuracy (HHAR)} & \textbf{94.79}&	91.5&	78.55&	92.67&	92&	91.33&	92.56&	92.51&	92.66&	92.83 \\
 \hline
 \textbf{F1 Score (HHAR)}& \textbf{0.792}& 0.671	&0.601&	0.671&	0.678&	0.677&	0.712&	0.702&	0.704&	0.708\\
 \hline
 \textbf{NLL (HHAR)} & \textbf{2.558}&	42.14&	3.75&	30.14&	8.01&	5.58&	4.13&	23.54&	7.18&	4.36 \\
 \hline
  \hline
  \textbf{Accuracy (Opp.)} & \textbf{86.80}&	83.85&	80.03&	83.24&	82.81&	83.85&	84.63&	82.35&	82.99&	84.06 \\
 \hline
 \textbf{F1 Score (Opp.)}& \textbf{0.808}&	0.732&	0.701&	0.741&	0.742&	0.744&	0.765&	0.743&	0.746&	0.755\\
 \hline
 \textbf{NLL (Opp.)} & 2.39&	12.35&	4.00&	12.31&9.38&	8.38&	\textbf{2.28}&	9.21&	7.98&	3.60 \\
 \hline
  \hline
 \textbf{Accuracy (SC)} & \textbf{82.12}&	78.45&	71.15&	78.48&	80.02&	80.65&	81.01&	79.33&	80.45&	81.01 \\
 \hline
 \textbf{F1 Score (SC)}& \textbf{0.733} &0.622&	0.589&	0.624&	0.628&	0.633&	0.706&	0.596&	0.599&	0.675\\
 \hline
 \textbf{NLL (SC)} & \textbf{1.01}&2.11&	1.97&	1.44&	1.20&	1.13&	1.08&	1.44&	1.18&	1.08 \\
  \hline
  \hline
\end{tabular}}
\end{center}
% \vspace{-0.10in}
\end{table*}

 \begin{figure*}[t]
    \centering
    \includegraphics[width=2\columnwidth]{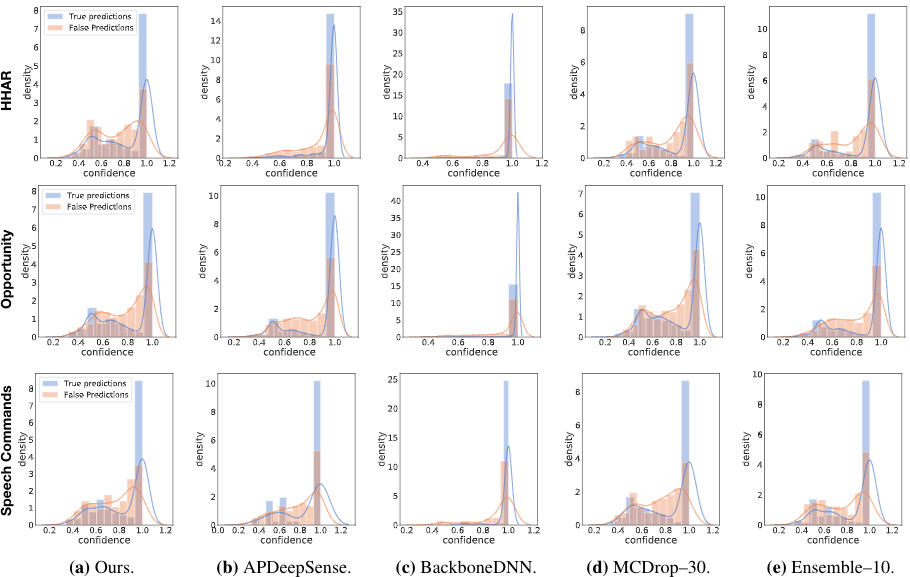}
    % \vspace{-0.15in}
    \caption{Density histogram of confidence measures for true (correct) and false (incorrect) predictions. A distribution skewed towards the right (near 1.0 on the x-axis) indicates the model has higher confidence in predictions than the distribution skewed towards left.
    [The density histogram is a histogram with area normalized to one. Plots are overlaid with kernel density curves for better readability.]
    }
    \label{fig:confidence_all}
    % \vspace{-0.15in}
\end{figure*}

\begin{figure*}[t]
    \centering
    \begin{subfigure}[t]{0.33\linewidth}
        \centering
        \includegraphics[trim=0 0 0 0,clip,width=\linewidth]{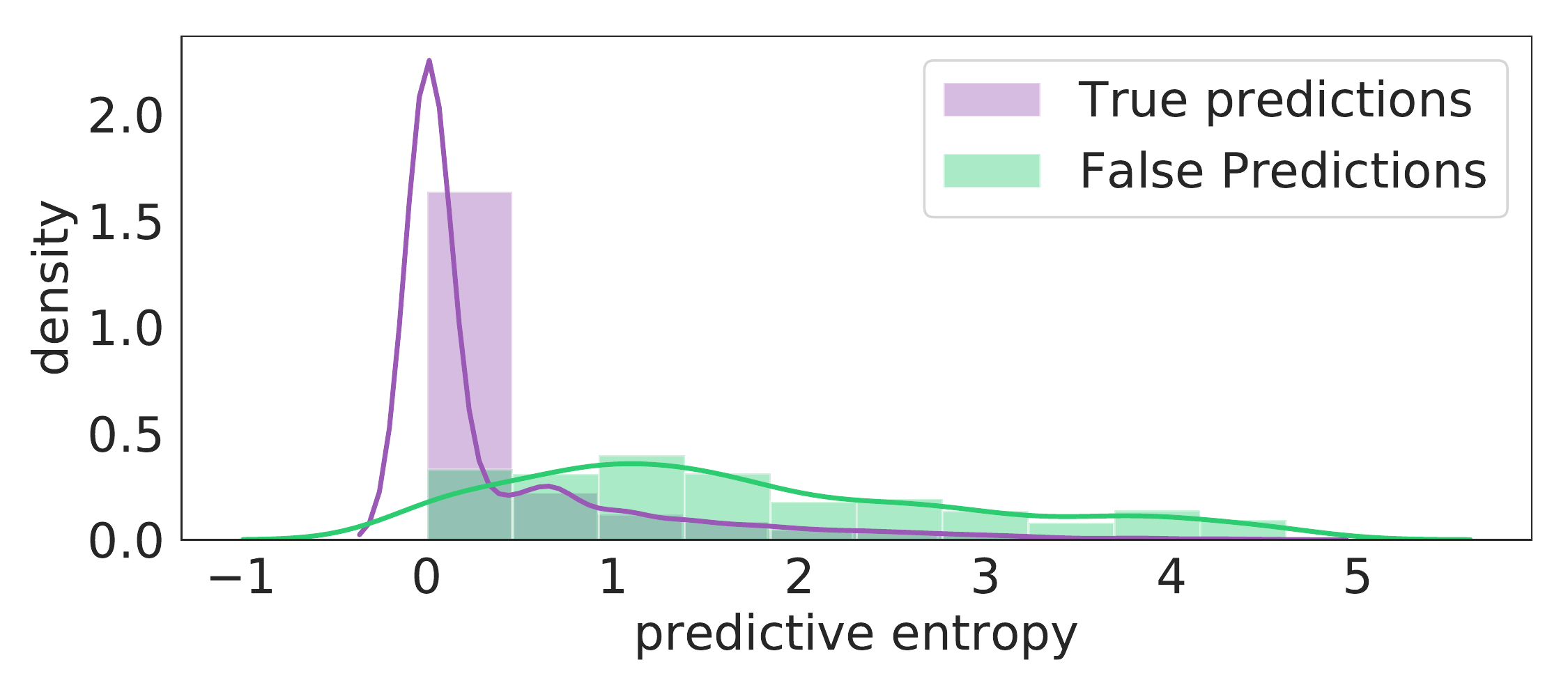}
        \label{fig:hhar_entropy_ours}
    \end{subfigure}
    % \hspace{0.02\textwidth}
    \begin{subfigure}[t]{0.33\textwidth}
        \centering
        \includegraphics[trim=0 0 0 0,clip,width=\textwidth]{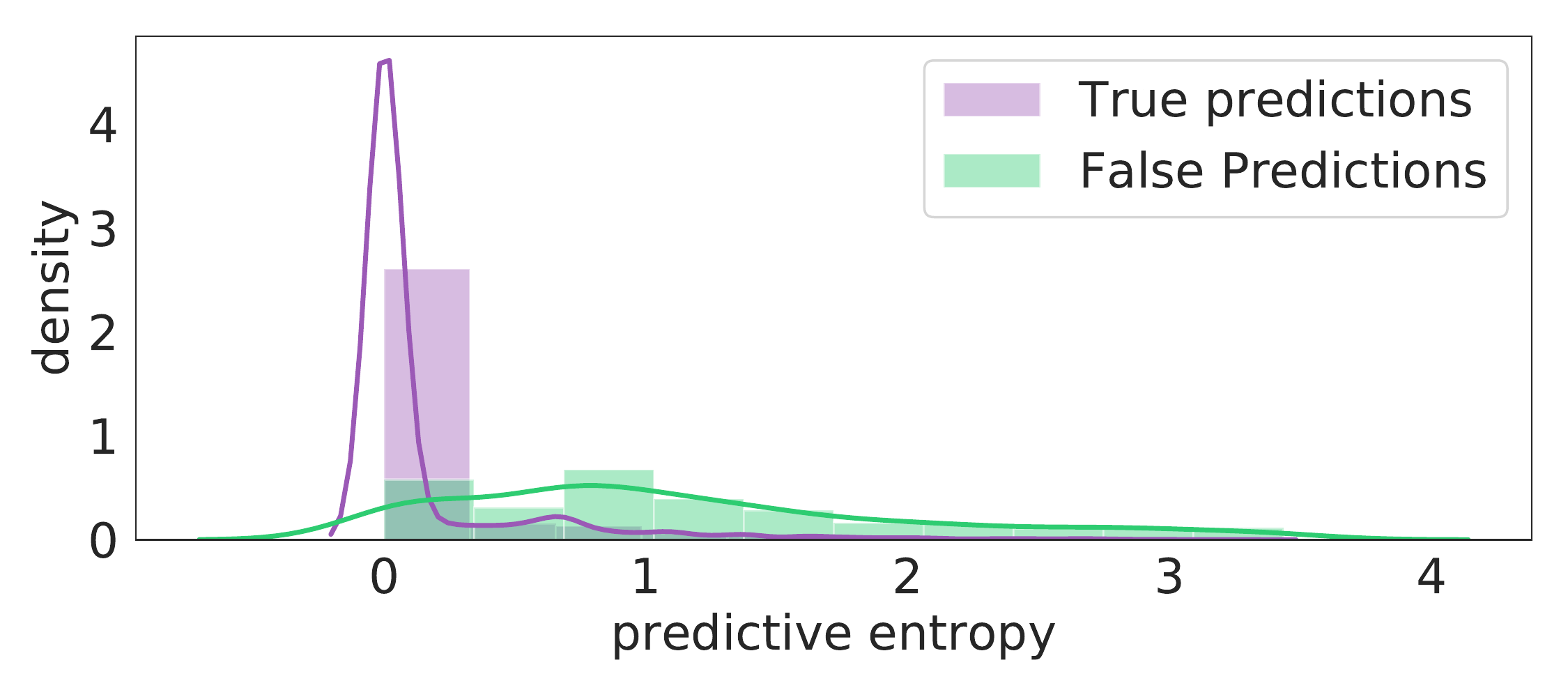}
        \label{fig:hhar_entropy_mc30}
          \vspace{-0.25in}
        \caption{Heterogeneous Human Activity Recognition}
    \end{subfigure}
    \begin{subfigure}[t]{0.33\textwidth}
        \centering
        \includegraphics[trim=0 0 0 0,clip,width=\textwidth]{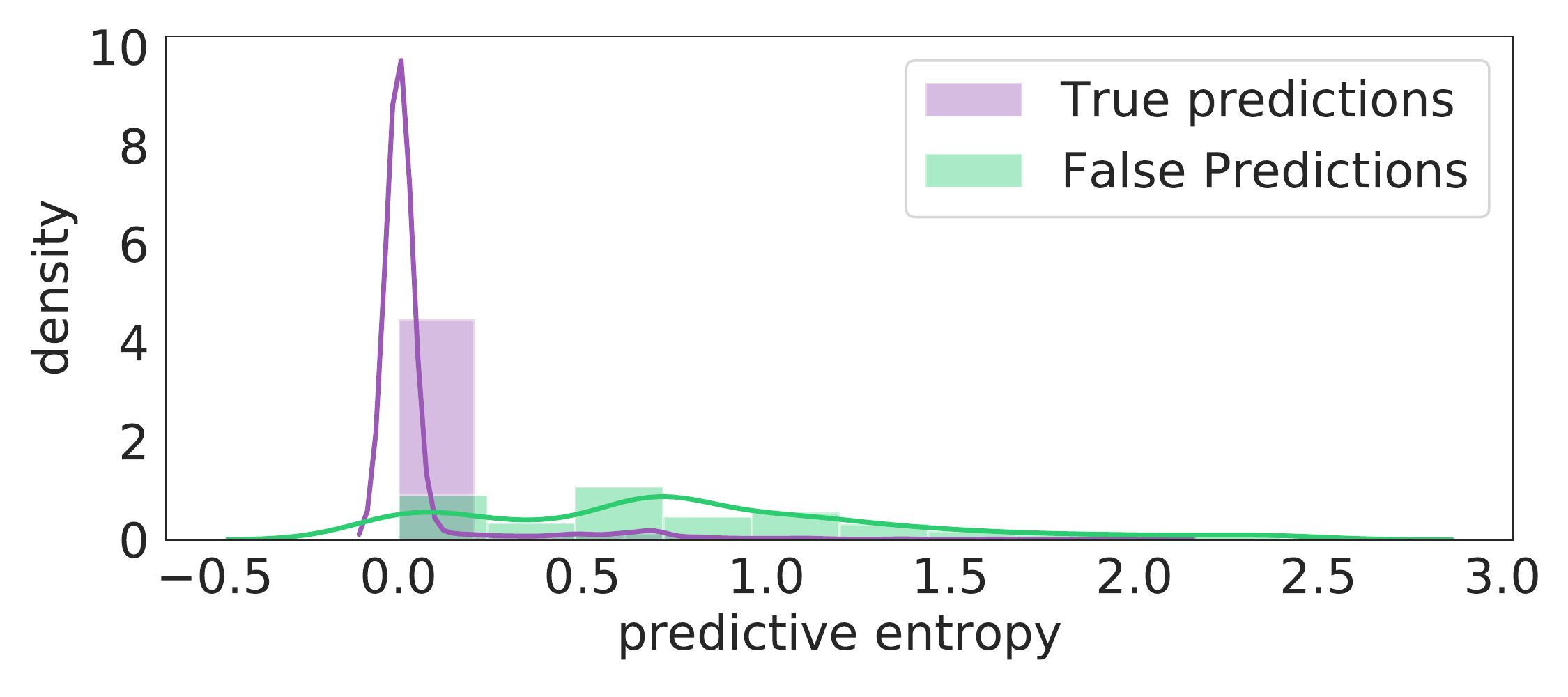}
        \label{fig:hhar_entropy_en10}
    \end{subfigure}
    \begin{subfigure}[t]{0.33\textwidth}
        \centering
        \includegraphics[trim=0 0 0 0,clip,width=\textwidth]{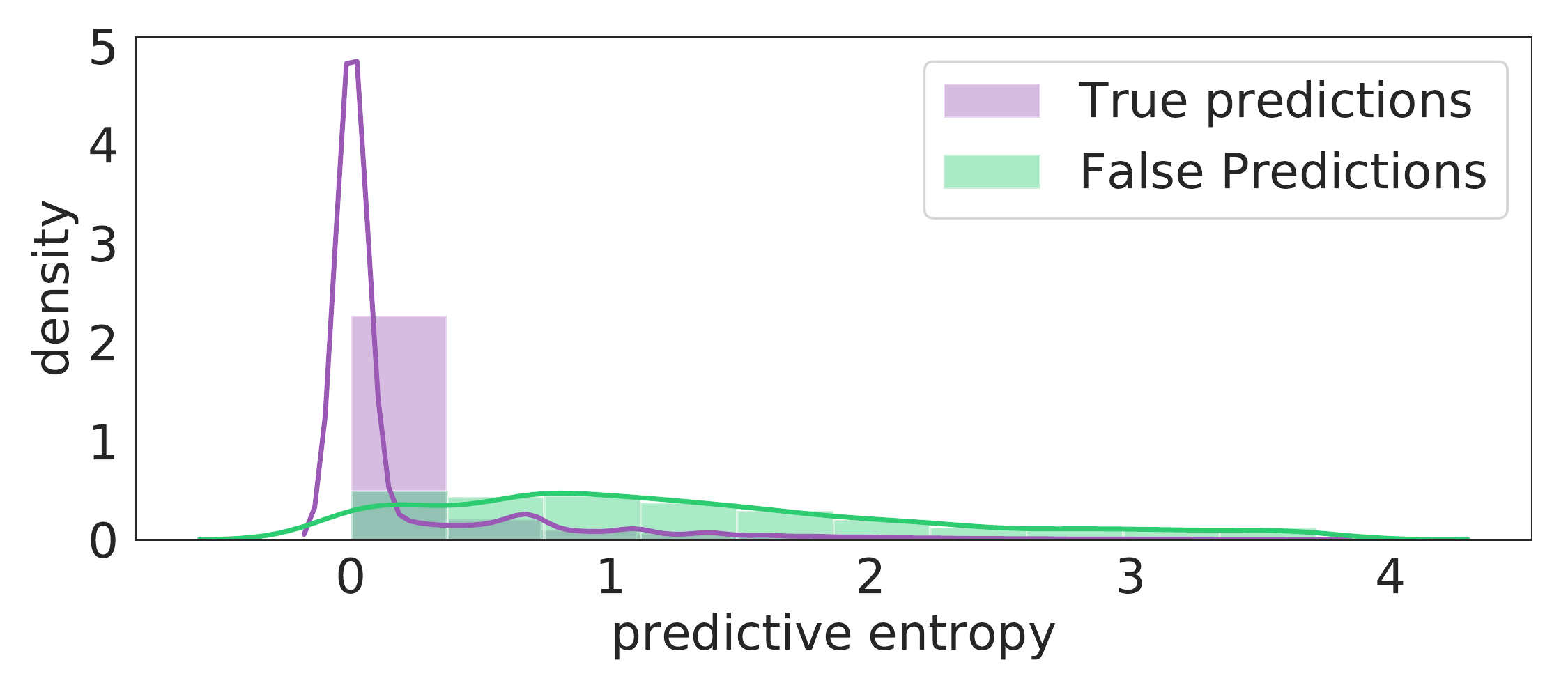}
        \label{fig:opportunity_entropy_ours}
    \end{subfigure}
    \begin{subfigure}[t]{0.33\textwidth}
        \centering
        \includegraphics[trim=0 0 0 0,clip,width=\textwidth]{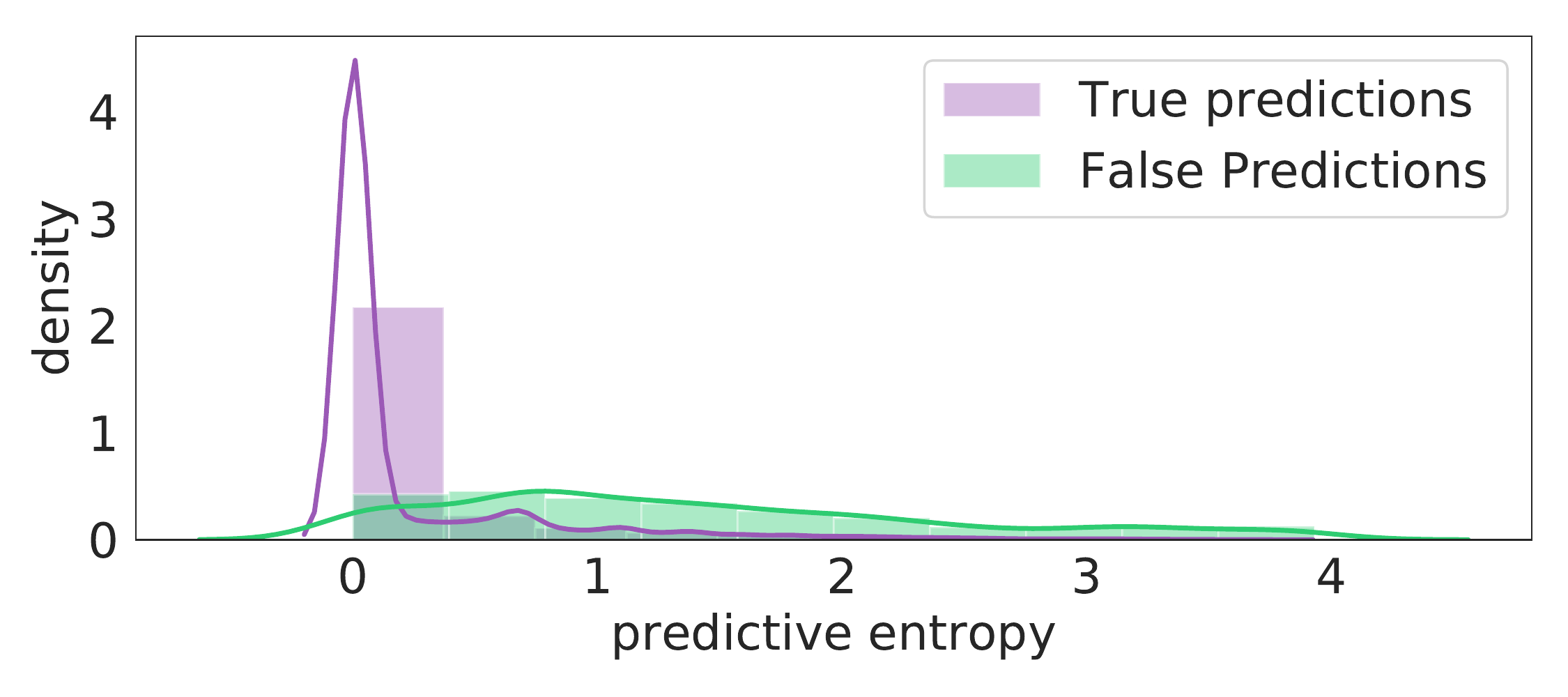}
        \label{fig:opportunity_entropy_mc30}
        \vspace{-0.25in}
        \caption{Opportunity}
    \end{subfigure}
    \begin{subfigure}[t]{0.33\textwidth}
        \centering
        \includegraphics[trim=0 0 0 0,clip,width=\textwidth]{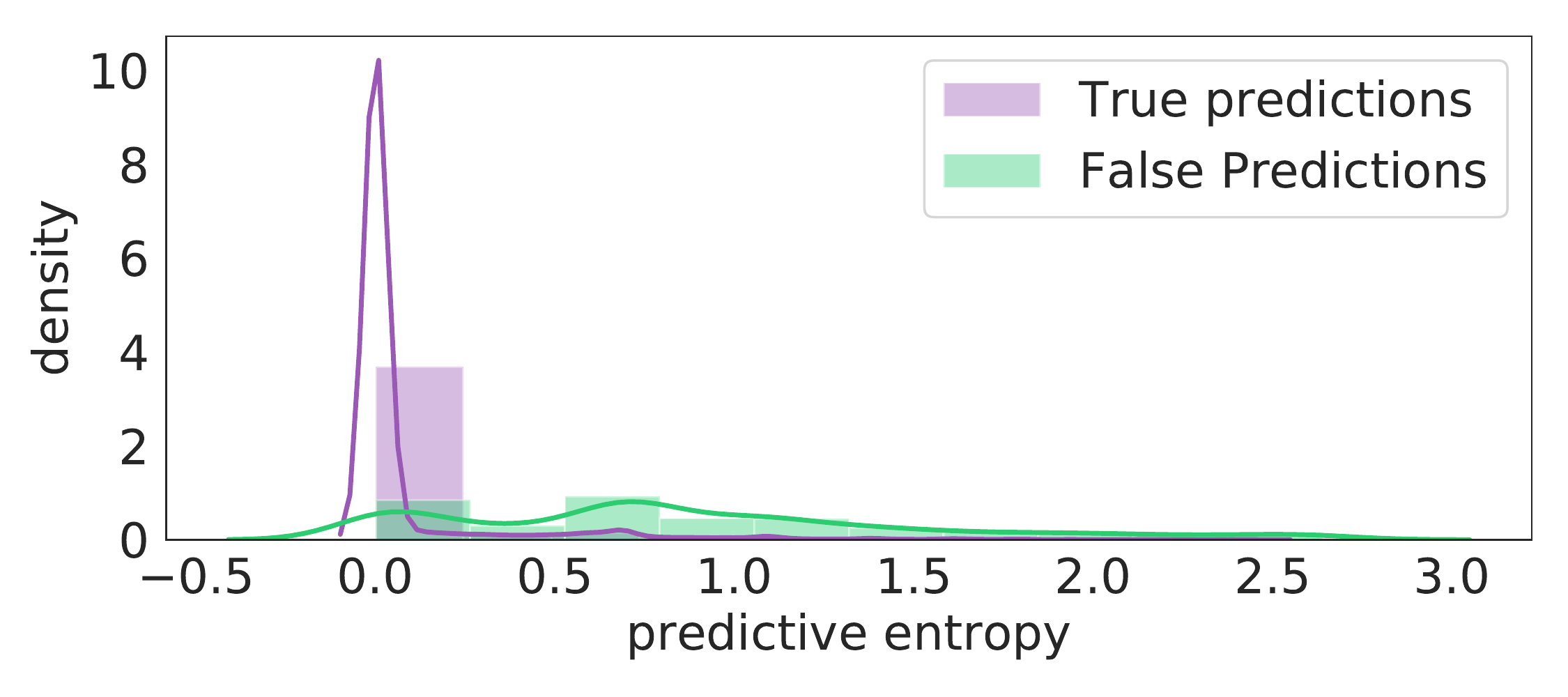}
        \label{fig:opportunity_entropy_en10}
    \end{subfigure}
    \begin{subfigure}[t]{0.33\textwidth}
        \centering
        \includegraphics[width=\textwidth]{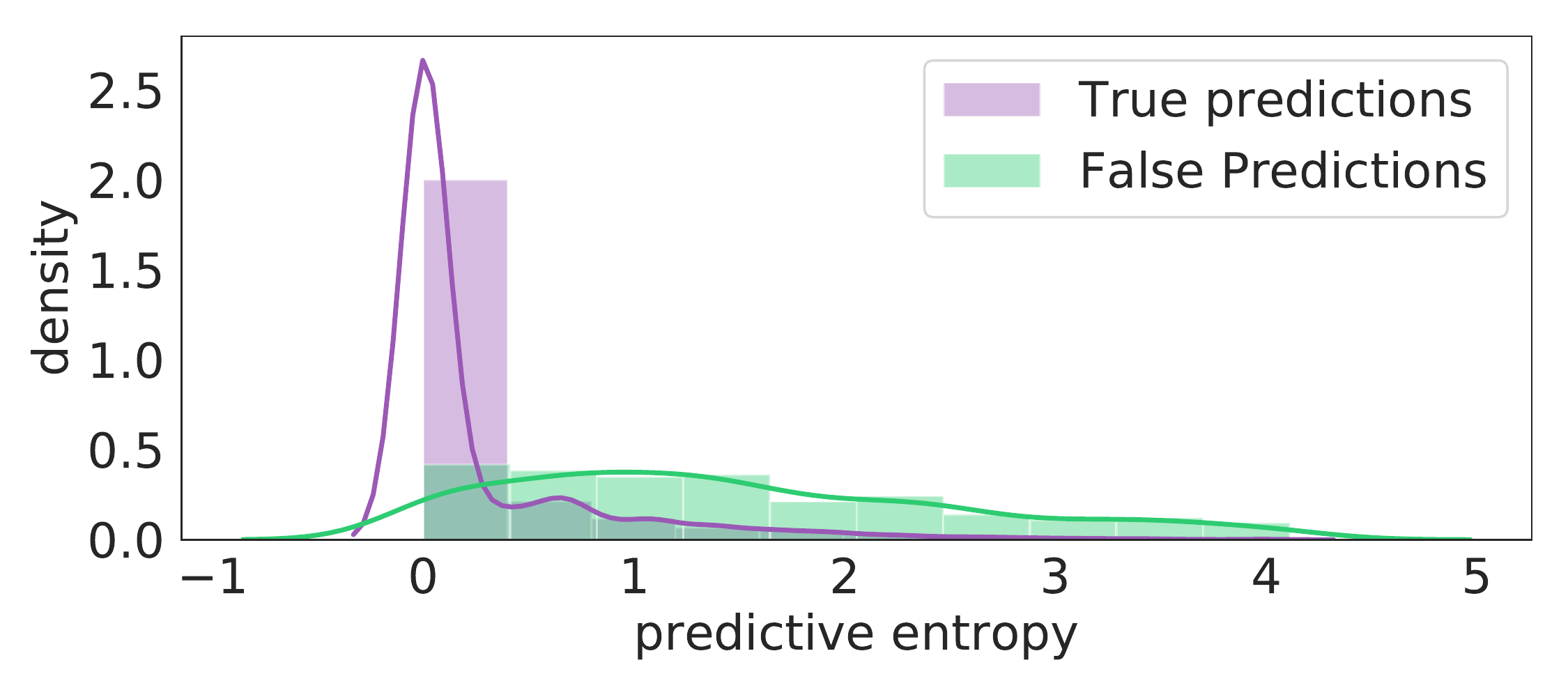}
        \label{fig:speech_entropy_ours}
    \end{subfigure}
    \begin{subfigure}[t]{0.33\textwidth}
        \centering
        \includegraphics[width=\textwidth]{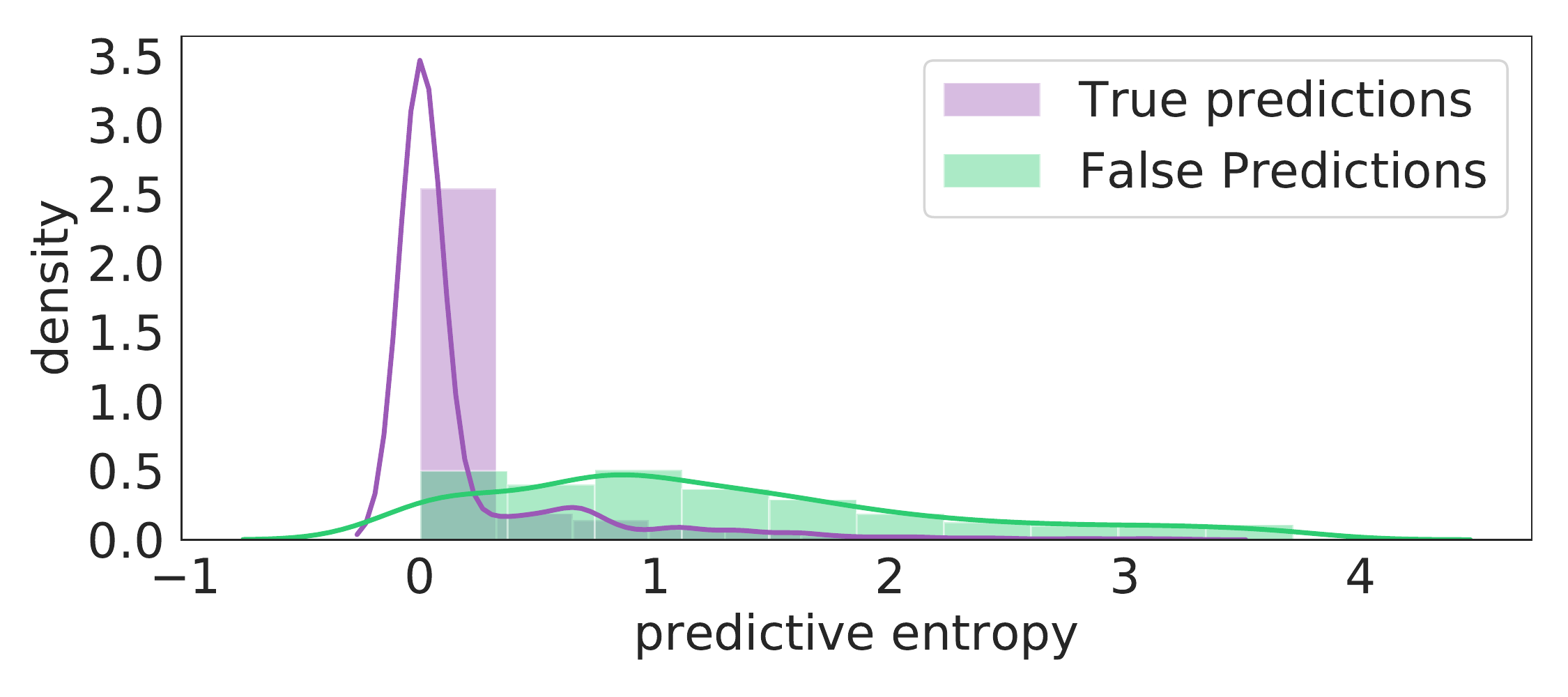}
  \label{fig:speech_entropy_mc30}
   \vspace{-0.25in}
    \caption{Speech Commands}
    \end{subfigure}
    \begin{subfigure}[t]{0.33\textwidth}
        \centering
        \includegraphics[width=\textwidth]{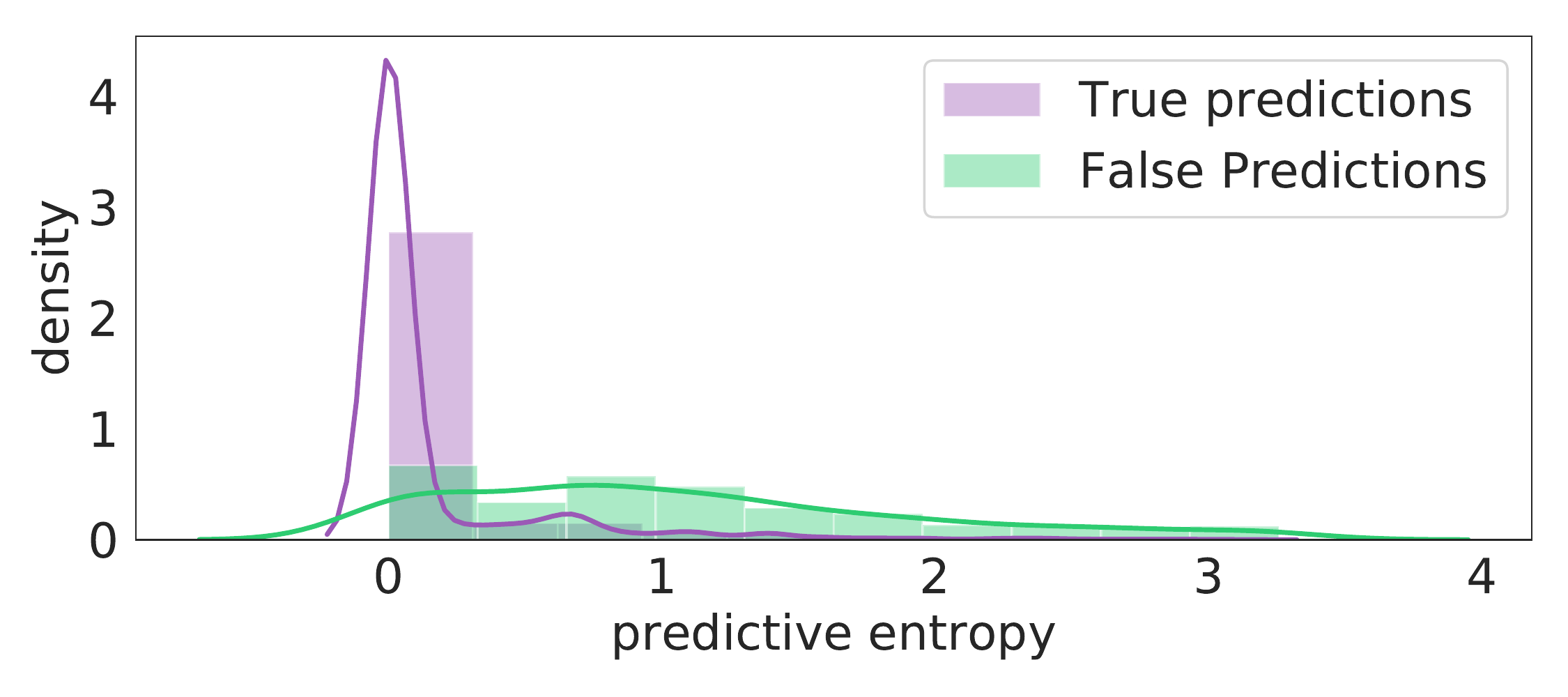}
        \label{fig:speech_entropy_en10}
    \end{subfigure}
    \vspace{-0.10in}
    \caption{Density histogram of predictive entropy for true (correct) and false (incorrect) predictions. Predictive entropy $H(\textbf{y}|\textbf{x}, D)$ which captures the average amount of information contained in the predictive distribution. To reliably capture the predictive uncertainty, we aim for a predictive entropy which is low for true and high for false inferences. [The density histogram is a histogram with area normalized to one. Plots are overlaid with kernel density curves for better readability.]
    }
    \label{fig:entropy_all}
    % \vspace{-0.15in}
\end{figure*}

\subsection{Baseline algorithms}
We tested our proposed method on the three datasets described above and compared the outcome with other four state-of-the-art approaches: conventional DNN, Monte Carlo Dropout (MCDrop)~\cite{gal2016dropout},  ApDeepSense~\cite{yao2018apdeepsense} and Deep Ensembles~\cite{lakshminarayanan2017simple}.

\textbf{BackboneDNN} is a conventional deep neural network. 
In order to show the benefits of a stochastic approach, we need to compare it to the traditional deep learning network. 
This network is used as the non-Bayesian baseline. As mentioned before, our technique and MCDrop rely on an already trained network, therefore, this network is the one we consider as the already trained network we refer to.

\textbf{Monte Carlo Dropout (MCDrop)} is based on Monte Carlo sampling and, runs the neural network multiple times to generate the uncertainty estimation. Hence, we use MCDrop-\textit{k} to represent MCDrop with \textit{k} (3, 5, 10, and 30) runs (forward passes). This approach, like ours, assumes that the model has already been trained with dropout. For this baseline, we keep dropout activated during inference too.

\textbf{ApDeepSense} is an algorithm that enables fully-connected NNs to provide uncertainty estimations during inference. This technique too uses dropout to perform the basic operations in the FC layers. Compared to our method, ApDeepSense works only with MLPs and considers mainly regression tasks where the uncertainty is represented by the variance of the distribution. Therefore, it does not translate very well to classification tasks. For this baseline, as suggested in the original proposal~\cite{yao2018apdeepsense}, we use a 5-layer neural network composed of fully connected layers with 512 hidden dimensions and ReLU activation function.

\textbf{Deep Ensembles} rely on providing uncertainty estimations by training and running an ensemble  of models (multiple networks). Although this baseline requires retraining, we include it to illustrate the upper bound of the uncertainty estimation quality that can be accomplished with retraining. Ensembles are created by training the models with random initialization~\cite{lakshminarayanan2017simple}. To achieve this, we use the Backbone architecture (not its trained model) with random initialization for each model. We use DeepEnsemble-\textit{k} to represent an ensemble of \textit{k} (3, 5, and 10) individual NNs.

\subsection{Quantitative Evaluations}
In this section, we present the results in terms of accuracy, F1 score and negative log-likelihood (NLL). The prediction accuracy expresses the correlation between the prediction of the deep neural network and the actual value, instead the F1 score is the weighted average of precision and recall. NLL measures the correspondence between the ground truth values and their predicted distributions. Lower NLL means higher correspondence.

% \vspace{-0.10in}
{
\begin{equation}
 \scriptstyle NLL\left(\mu, \sigma^{2} ; x_{1}, \ldots, x_{n}\right)=\frac{n}{2} \ln (2 \pi)+\frac{n}{2} \ln \left(\sigma^{2}\right)+\frac{1}{2 \sigma^{2}} \sum_{i=1}^{n}\left(x_{j}-\mu\right)^{2}   
\end{equation}
}

In addition to the aforementioned metrics, we consider the confidence and predictive entropy as measure of uncertainty. The confidence metric gives a better understanding on how the model behaves during inference. Ideally we want to achieve high confidence for correct predictions and low confidence for incorrect ones. In classification tasks, it is considered as the confidence given from the softmax. In conventional DNN and APDeepsense, it is measured as the result based on only one softmax operation. Instead in the other baselines, including ours, it is the mean of categorical predictive distribution.
To evaluate the predictive uncertainty we measure the predictive entropy $H(\textbf{y}|\textbf{x}, D)$ which captures the average amount of information contained in the predictive distribution.

% \vspace{-0.10in}
{
\begin{equation}
    \scriptstyle H(\textbf{y}|\textbf{x}, D) \scriptstyle = \scriptstyle - \sum_{i=0}^{K-1}p_{i\mu} * log p_{i\mu},
\end{equation}
}

where $p_{i\mu}$ is the predictive mean probability of the $i^{th}$ class from $T$ Monte Carlo samples in the case of MCDrop, the $K$ model predictions in the case of Deep Ensembles, and finally, from the $j$ unaries sampled from the output distribution in our approach. To reliably capture the predictive uncertainty, we aim for a  predictive entropy which is low for true and high for false inferences.

\begin{figure*}[t]
    \centering
    \begin{subfigure}[t]{0.33\linewidth}
        \centering
        \includegraphics[width=\linewidth]{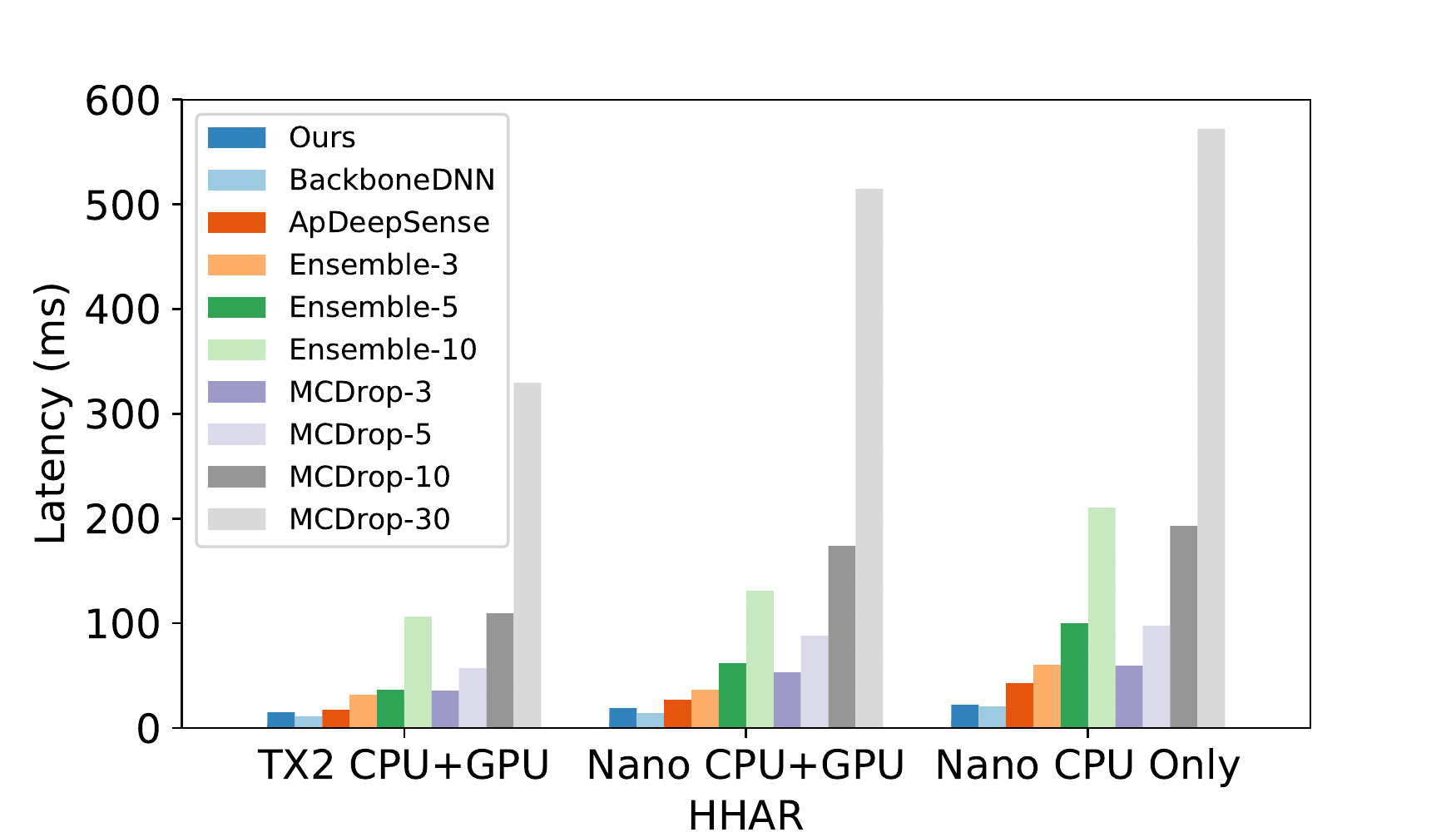}
        % \vspace{-2in}
        % \caption{HHAR}
        \label{fig:lathhar}
    \end{subfigure}
    % \hspace{0.02\textwidth}
    \begin{subfigure}[t]{0.33\textwidth}
        \centering
        \includegraphics[width=\textwidth]{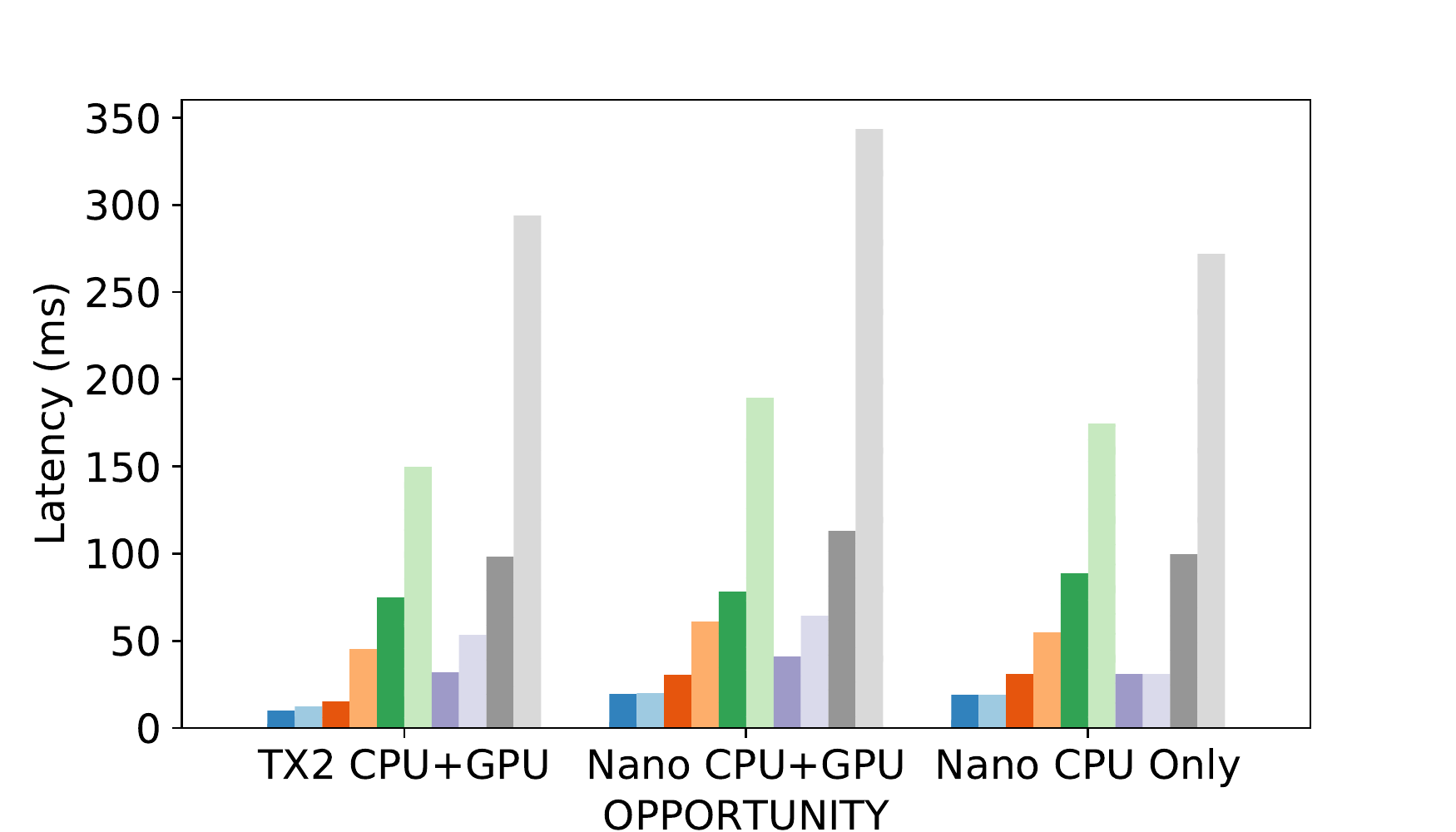}
        % \vspace{-2in}
        % \caption{Opportunity}
        \label{fig:latopp}
    \end{subfigure}
    \begin{subfigure}[t]{0.33\textwidth}
        \centering
        \includegraphics[width=\textwidth]{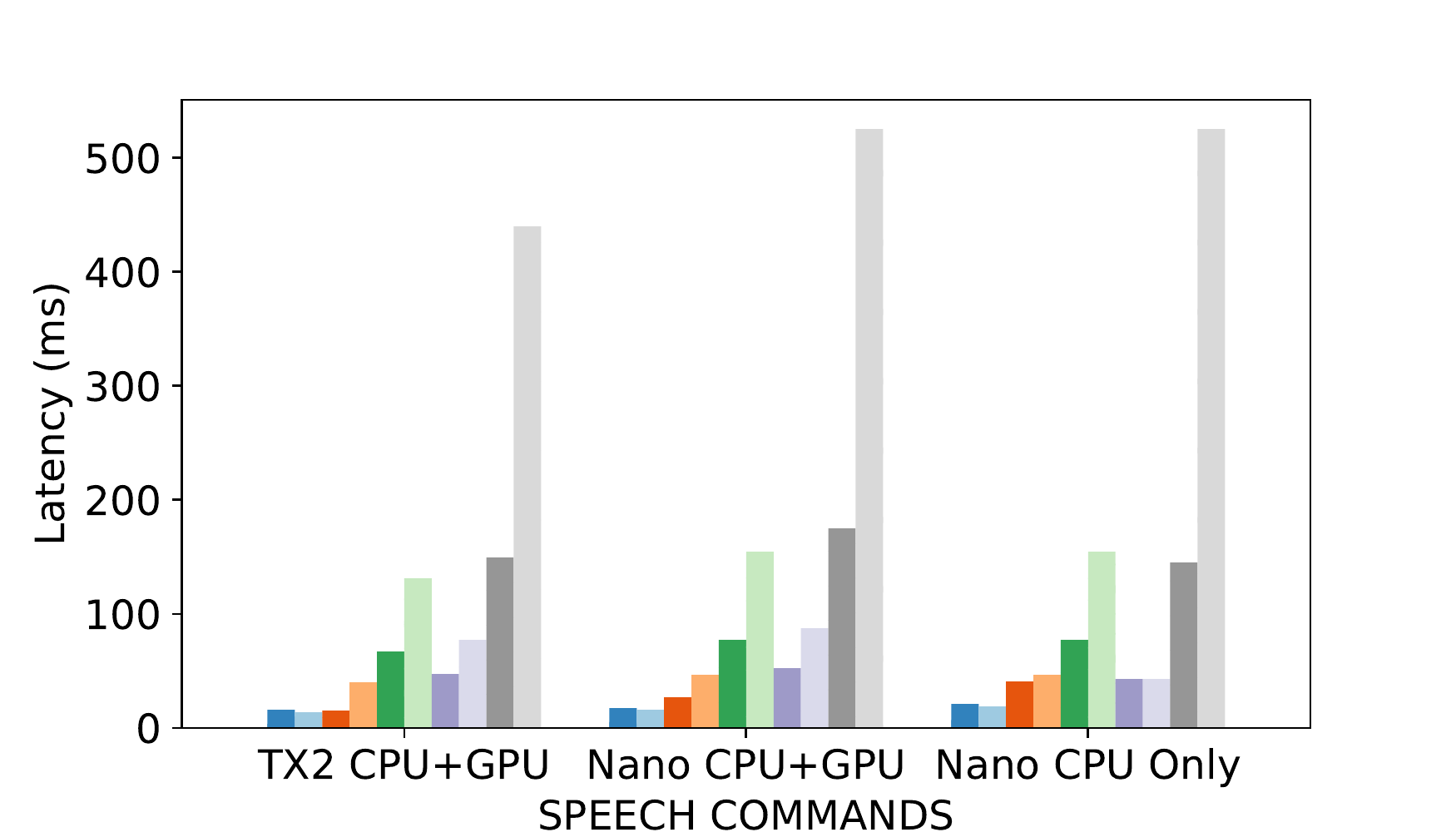}
        % \vspace{-2in}
        % \caption{SC}
        \label{fig:latsc}
    \end{subfigure}
    \begin{subfigure}[t]{0.33\textwidth}
        \centering
        \includegraphics[width=\textwidth]{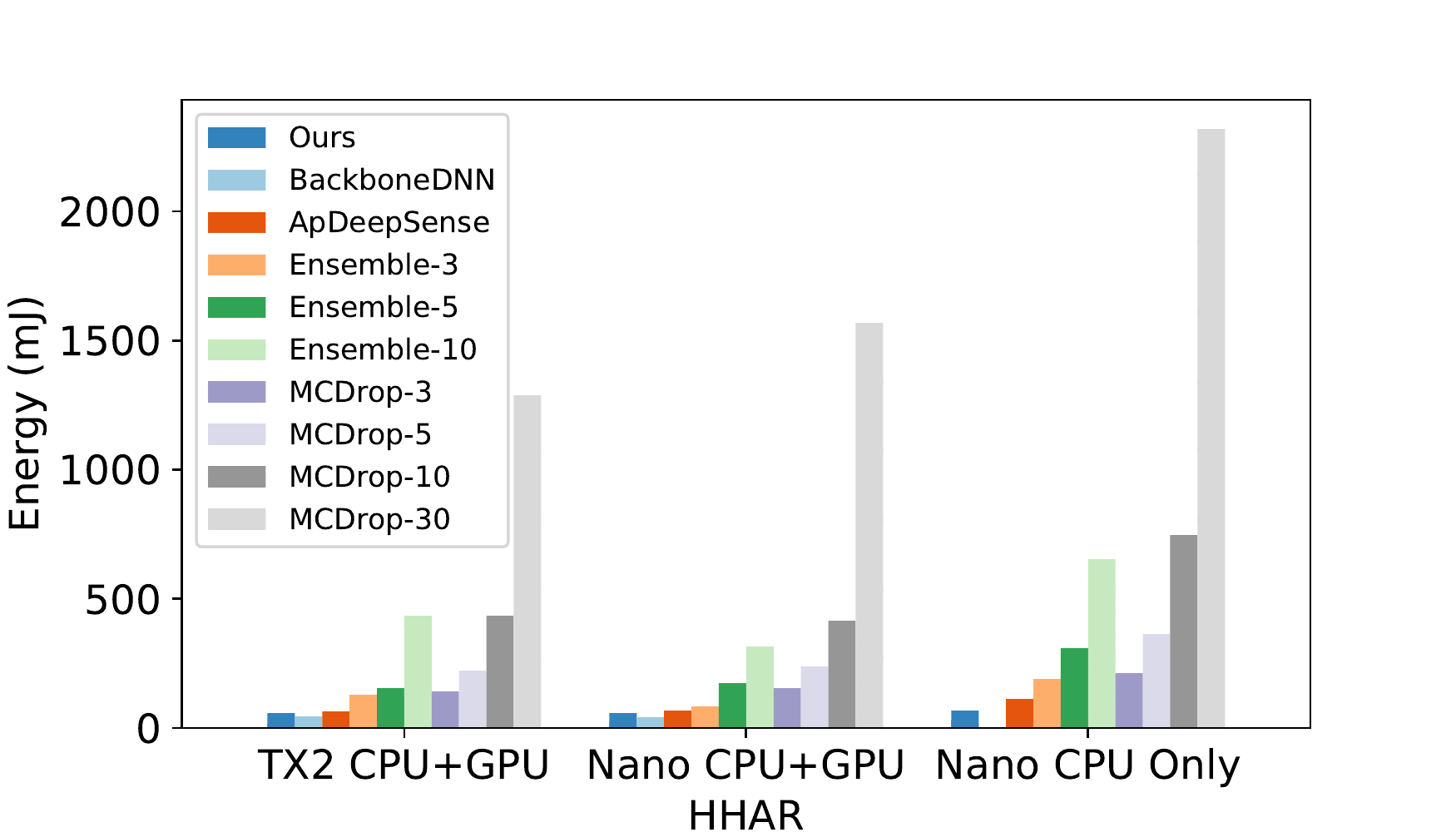}
        % \caption{HHAR}
        \label{fig:energyhhar}
    \end{subfigure}
    \begin{subfigure}[t]{0.33\textwidth}
        \centering
        \includegraphics[width=\textwidth]{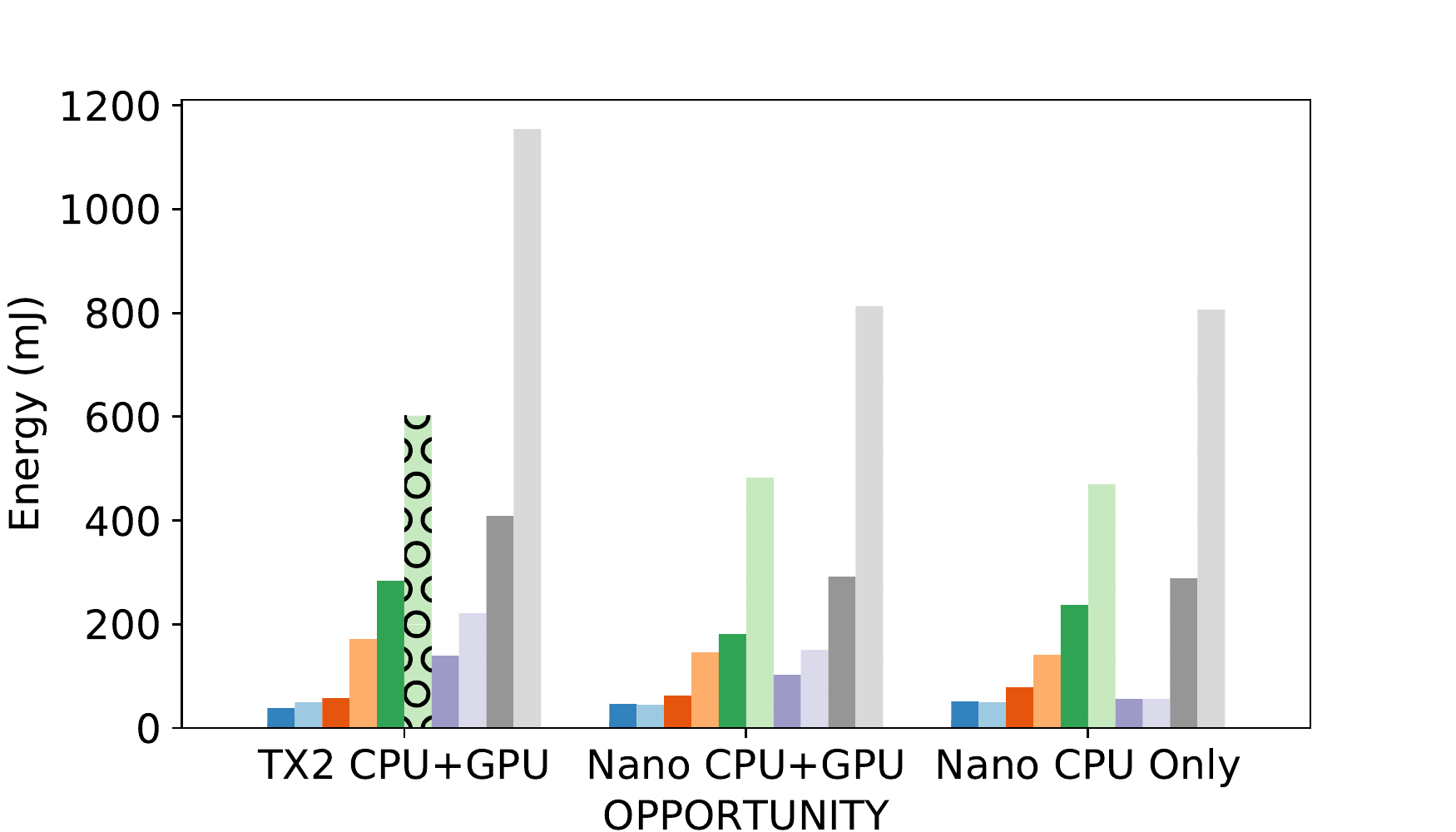}
        % \caption{Opportunity}
        \label{fig:energyopp}
    \end{subfigure}
    \begin{subfigure}[t]{0.33\textwidth}
        \centering
        \includegraphics[width=\textwidth]{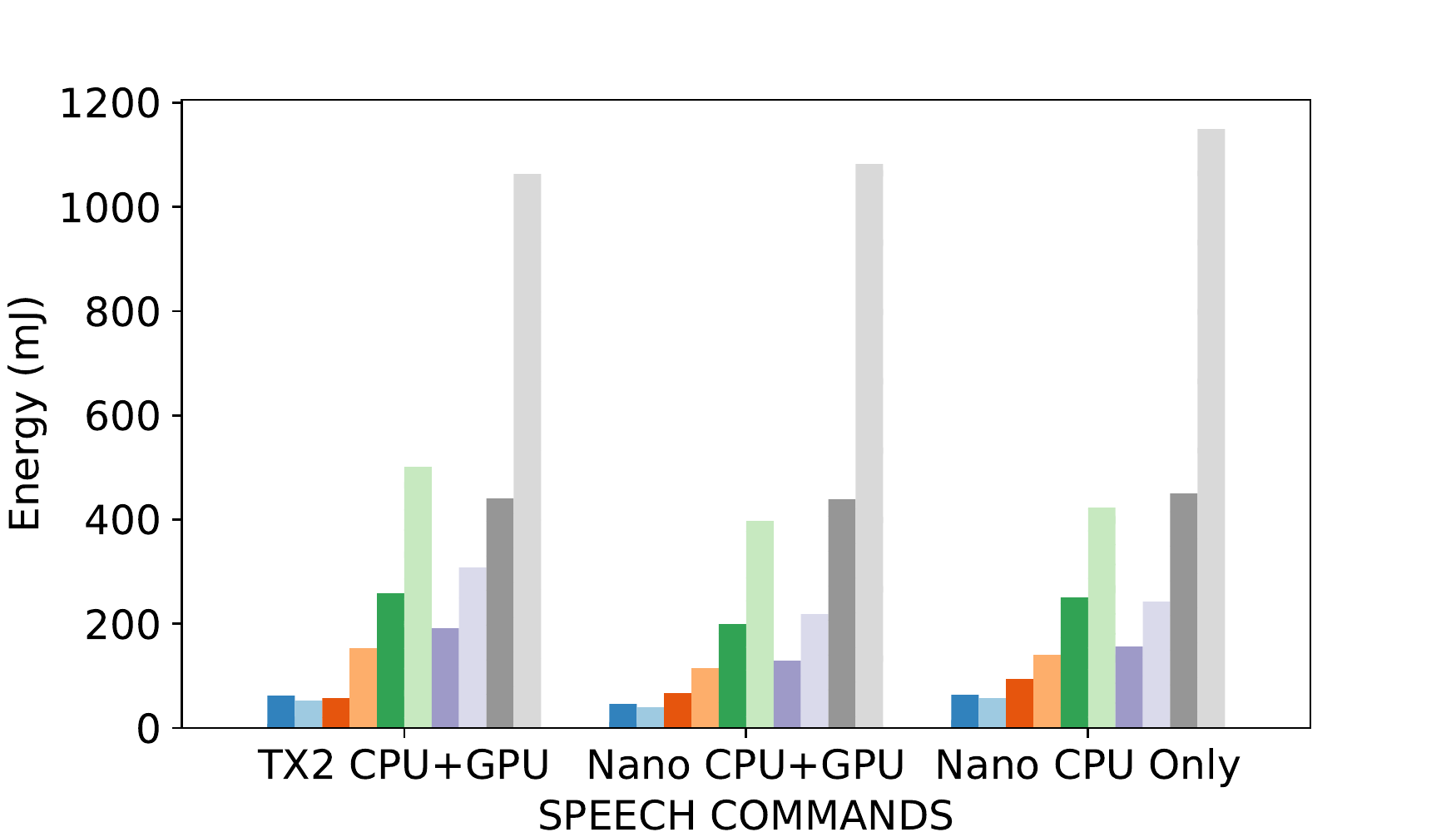}
        % \caption{SC}
        \label{fig:energysc}
    \end{subfigure}
    % \vspace{-0.10in}
    \caption{Inference time and Energy consumption on two edge platforms with different CPU and GPU capacity.  The latency indicates the average time it takes for the deep learning model to make a prediction on the provided sample. The energy consumed is expressed as \textit{power x time}.
    }
    \label{fig:latency}
    % \vspace{-0.15in}
\end{figure*}

\subsection{Embedded Edge Systems Setup}
To evaluate our framework's performance, we run the inference on two edge platforms,  NVIDIA Jetson TX2~\cite{jetson} and Nano~\cite{nano}, and measure the aforementioned metrics while computing the latency and energy consumption per testing sample. The latency indicates the average time it takes for the model to make a prediction on the provided sample. The energy consumed is expressed as \textit{power x time}. The TX2 is an embedded system-on-module and it is representative of today's embedded platforms with capabilities (256-core Pascal GPU, 2-core Denver + 4-core ARM A57 CPU, 8GB RAM, input \textasciitilde 19V) similar to high-end smartphones such as Samsung Galaxy 20 and OnePlus 7 Pro (Octa-core CPUs, Adreno 640/650 GPU, and 12GB RAM). The Nano, instead, has lower capabilities (128-core Maxwell GPU, 4-core ARM A57 CPU, 4GB RAM, input \textasciitilde 5V) and highlights the efficiency and feasibility of our approach (see Section~\ref{energy}) to run on more constraint embedded platforms. Both platforms have CPU and GPU support for deep learning libraries (we use tensorflow 1.15.2).
For an evalulation catering for more resource-limited devices,
we evaluate our framework on the Jetson Nano with CPU only enabled (with all the datasets). This validates the applicability to these kind of limited platforms.

\subsection{Model Estimation Performance}

Table~\ref{tab:acc1} shows the accuracy, F1 scores and negative log-likelihood obtained for the three
% HHAR, Opportunity and Speech Command 
datasets.  Our approach is the best performing across them: we can provide accurate predictions with high-quality uncertainty estimates. 
We achieve higher accuracy compared to the other methods, especially ApDeepSense, because we employ CNNs instead of MLPs.
% Lower NLL means that better uncertainty estimates are obtained through layerwise approximations. 
% The robustness of MCDrop--k increases when $k$ increases. With $k$ increases also the number of Monte Carlo samples which provide better uncertainty estimates but a higher resource footprint and latency.
MCDrop does achieve accuracy similar to our method, however, it takes 30 runs to reach a NLL close to the one of our method, which  makes it very power hungry  and not suitable for resource-constrained devices as highlighted in Section~\ref{energy}. The similar argument holds for Ensemble--10. An ensemble not only requires running multiple models in order to get the predictive uncertainty, but it also demands for keeping them in memory. In case of many embedded devices this would involve some kind of scheduling for memory allocation to perform them all. In order to apply this technique, it is required to train multiple models, therefore, it cannot be performed on already trained networks.

To have a closer look at what happens when we compare the best performing approaches at a more fine-grained level, we present the results on confidence and predictive entropy in the following density histograms.
We notice how the trust concern in the overconfident predictions of the conventional DNN BackboneDNN is valid for all three datasets (see Figure~\ref{fig:confidence_all}). 
Even if the model has high accuracy, it loses its credibility when the confidence is so high for false predictions. We see a peak near higher confidence values for true predictions. However, most importantly, we want lower confidence values for false predictions. Our approach conforms to this as the best performing approach especially in the HHAR dataset.
ApDeepSense performs worse, as expected, given its limitations in relying only on MLPs and moreover it is not enabled to work well for classification. 

In Figure~\ref{fig:entropy_all}, the density histograms illustrate the predictive entropy as uncertainty estimate. Predictive entropy embodies the average amount of information in the predictive distribution. We compare our approach versus MCDrop--30 and Ensemble--10, the best performing so far and best SOTA baselines. 
Although the three techniques can provide the uncertainty measure, ours performs better than Ensemble--10 on all three datasets and better than MCDrop--30 on the HHAR. Although, our results are similar to MCDrop--30 on the other two datasets, we obtain the illustrated uncertainty estimates in one single forward pass with great computational advantage and only a slight computation overhead compared to vanilla DNNs (see Section~\ref{energy}).

\subsection{Latency and Energy Consumption}
\label{energy}
Figure~\ref{fig:latency} shows  the latency (inference time) and energy consumption for all datasets. The experiments are conducted on an Nvidia Jetson TX2 and Nano. For fairness, we measure only the time needed to pass a sample and not consider the time needed to upload the model. In most of the considered baselines including ours, the model is uploaded just once and kept in memory. However, for ensembles, this might be different depending on the capacity and scenario, therefore we decide not to add that time and computation to the results.
As we can see, our approach adds only a slight (max 20\%) overhead over the conventional BackboneDNN, while being able to provide uncertainty estimates. Our latency is around 9-19ms per inference depending on the dataset and the edge platform. 
The latency of MCDrop is significantly worse and the time to perform inference increases with the number of runs (forward passes) being 20x times in the best scenario compared to our method and increasingly more in other cases (up to 28--fold). Similar trends can  be observed for Ensembles (2x - 8x times) as they require running multiple neural networks.

The energy consumption measurements show similar patterns. Ours is also at least (20\%) faster compared to one of the most recent approaches (ApDeepSense). In general, our method always requires less energy than all the other approaches and adds only a negligible or a tiny overhead (depending on the dataset) on the traditional DNN approach which does not provide uncertainty estimates.
Noticeably, our approach performs well on  Nano, highlighting the fact that the  applications can    harness the utility of reliable predictions  on many modern mobile and embedded devices especially if latency could be slightly sacrificed (which is often the case for critical applications). Additionally,   our CPU only results on Nano demonstrates that   our  framework can run efficiently on resource-constrained devices that do not have a GPU.

As mentioned before, we want to make sure that our models can have a small footprint on these devices. For both embedded platforms, our investigations show that we add only a negligible runtime memory overhead (max 5\%) compared to the vanilla deep learning model, while improving on the MLP baseline by 30\%. Deep ensembles start heavily using the memory swap on the Nano when passing 5 ensembles, therefore, there is more need for memory sharing mechanisms. MCDrop does not contribute extra to the memory but, of course, this technique relies on a lengthy computation time as seen in all the results in this section.

%% file: discussion.tex
\section{Discussion and Conclusions}
\label{discussion}
  We have introduced a framework  able to provide uncertainty estimates through layerwise distribution approximations using only a single forward pass. It provides a simple and system efficient solution that empowers convolutional neural networks to generate  uncertainty estimations during inference for classification tasks.  We evaluated our approach on multiple mobile systems datasets and showed that it significantly outperforms state-of-the-art baselines in terms of computation, latency and energy expenditure while providing reliable uncertainty estimates and accurate predictions.  
  
  There are many interesting avenues we want to pursue further in the future. 
 Our approach could be extended to recurrent neural networks. However, these alterations require additional effort in providing the right mathematical foundations and test its feasibility on real-life datasets. 
 
A key advantage of our framework is the fact that we model each layer to output predictive distributions.
 This can be a very flexible instrument for many applications, e.g., early prediction models, a class of conditional computation models that exit once a criterion (e.g., sufficient accuracy and low uncertainty) is satisfied at early layers. Such models can be very useful in intermittent learning systems~\cite{lee2019neuro, montanari2020eperceptive}  which are powered by harvested energy.
 
\new{Our approach, based on Gaussian approximations to the internal statistics of the network, is a feasible solution to providing uncertainty estimates on edge devices. These platforms cannot afford an increase in latency, memory or energy due to additional forward passes but that would benefit from uncertainty quantification. It is also a powerful solution considering that the approach does not require re-training or fine-tuning. }
 \new{This approximation
could be improved to consider the fact that
 the outputs of non-linear activations naturally yield skewed distributions with values possibly in a limited subset of the domain, which are not perfectly Gaussian distributed. Future work, therefore, could consider other approximations that minimize this discrepancy while simultaneously yielding uncertainty estimates for existing neural networks without increased operational costs, as is the case with our approach.}

To conclude, uncertainty estimations bring the much required element of interpretability and reasoning on the predictions made by neural network models. Such estimates are vital in the area of mobile and embedded systems as these systems deal with different kind of uncertainties. We have offered an avenue to provide them cheaply on these platforms while maintaining the needed level of performance.

\section{Acknowledgements}
This work is supported by Nokia Bell Labs through their donation for the Centre of Mobile,
Wearable Systems and Augmented Intelligence to the University of Cambridge. The authors
declare that they have no conflict of interest with respect to the publication of this work.